\documentclass{ifacconf}

\usepackage{graphicx}      
\usepackage{natbib}        
\usepackage[caption=false,font=footnotesize]{subfig}
\begin{document}
	
	\begin{frontmatter}
		
		\title{Results from the Robocademy ITN: Autonomy, Disturbance Rejection and
			Perception for Advanced Marine Robotics\thanksref{footnoteinfo}} 
		
		\thanks[footnoteinfo]{This work has been partially supported by the
			FP7-PEOPLE-2013-ITN project ROBOCADEMY (Ref 608096) funded by the European
			Commission.}
				
		\author[OSL]{Matias Valdenegro-Toro}
		\author[OSL]{Mariela De Lucas Alvarez}
	    \author[OSL]{Mariia Dmitrieva}
		\author[DFKI]{Bilal Wehbe}
		\author[NOC]{Georgios Salavasidis}
		\author[NTUA]{Shahab Heshmati-Alamdari}
		\author[TTU]{Juan F. Fuentes-P{\'{e}}rez}
		\author[Atlas]{Veronika Yordanova}
		\author[UdG]{Klemen Isteni\v{c}}
		\author[SeeByte]{Thomas Guerneve}
						
		\address[OSL]{Ocean Systems Lab, Heriot-Watt University, 
			EH14 4AS Edinburgh, Scotland, UK. (e-mail:  $\{$m.valdenegro,m.delucasalvarez$\}$@hw.ac.uk).}
		
		\address[DFKI]{DFKI - Robotic Innovation Center, Bremen, Germany
			(e-mail: \{bilal.wehbe, samy.marcelo\_nascimento\}@dfki.de)}
		
		\address[NOC]{Marine Autonomous and Robotics Systems, National Oceanography Centre, Southampton, SO14 1NJ, UK (e-mail: $\{$geosal,abp$\}$@noc.ac.uk)}
		
		\address[NTUA]{Control Systems Lab, Department of Mechanical Engineering, National Technical University of Athens, 9 Heroon Polytechniou Street, Zografou 15780. \\ E-mail: \{shahab,karrasg,kkyria\}@mail.ntua.gr}
		
		\address[TTU]{Centre for Biorobotics, Tallin University of Technology, Tallinn, Estonia, (e-mail: juan.fuentes@ttu.ee)}
		
		\address[Atlas]{Atlas Elektronik GmbH , Sebaldsbruecker Heerstrasse 235, Bremen, Germany.}
		
		\address[UdG]{Computer Vision and Robotics Institute (VICOROB), University of Girona, C\textbackslash Pic de Peguera, 13, 17003 Girona, Spain, (email: klemen.istenic@udg.edu)}
		
		\address[SeeByte]{SeeByte, Orchard Brae House, 30 Queensferry Rd, 
			EH4 2HS Edinburgh, Scotland, UK. (e-mail:  thomas.guerneve@seebyte.com).}
		
		\begin{abstract}                
			Marine and Underwater resources are important part of the economy of many countries. This requires significant financial resources into their construction and maintentance. Robotics is expected to fill this void, by automating and/or removing humans from hostile environments in order to easily perform maintenance tasks.
			The Robocademy Marie Sk\l{}odowska-Curie Initial Training Network was funded by the European Union's FP7 research program in order to train 13 Fellows into world-leading researchers in Marine and Underwater Robotics.
			The fellows developed guided research into three areas of key importance: Autonomy, Disturbance Rejection, and Perception.
			This paper presents a summary of the fellows' research in the three action lines. 71 scientific publications were the primary result of this project, with many other publications currently in the pipeline. Most of the fellows have found employment in Europe, which shows the high demand for this kind of experts. We believe the results from this project are already having an impact in the marine robotics industry, as key technologies are being adopted already.
		\end{abstract}
		
		
\end{frontmatter}

\section{Introduction}

Marine and Underwater environments are of high economic importance to many countries. For example, fisheries produce considerable food resources, and off-shore wind farms are key to producing clean energy. The increasing importance of these economic areas is also linked to higher maintenance and repair costs, specially for structures that are permanently underwater, and wind farms at sea.

Robotics and Automation has been proposed as a solution to control repair and maintenance costs, and to preserve the lives of workers, as operations at sea are quite dangerous. Particularly construction and maintenance operations underwater have to be performed by divers, with known risks to human life. Typically these risks are mitigated by insurance, which drives costs upward.	 

The field of Underwater Robotics has not seen the same kind of technological development than other robotics fields (like ground and aerial). This is caused by specifics of the underwater domain, like poor light visibility due to light absorption, making robots "sight" more difficult. Issues in perception translate into difficulties to perform manipulation. Acoustic sensing is a popular choice for underwater perception, as sound is not attenuated by water and can travel great distances, but the interpretation of acoustic sensor reading is difficult due to reflections, noise and unexpected interactions with the environment.

Controlling a underwater robot is also quite difficult, as there is a high degree of interaction with the medium (the water column), and unexpected underwater currents can move the robot without it being aware. Actuation is not trivial, as propellers are typically used which can fail due to bio-fouling. Any underwater vehicle design should be aware of these issues and consider methods to mitigate them. The autonomy level of underwater robots is therefore quite low, and there is great interest from marine-related industries (like oil and gas, renewables, and fisheries) to improve the autonomy of underwater robots in order to protect investments and reduce human risks.

It is in this context that the European Academy on Marine and Underwater Robotics (Robocademy) is born. Robocademy is structured as a Marie Sk\l{}odowska-Curie Initial Training Network (ITN) and was funded by the European Union's Research and Innovation funding programme FP7.
The key objective of the ITN is to train world-leading experts in Marine and Underwater robotics, supported by an ambitious training program both on their local institutions and across the network. In well-defined and well-tutored PhD research projects, the Robocademy fellows will push the state of the art in robust, reliable and autonomous underwater vehicles (AUVs).

The network consists of 10 partners across Europe, including Universities, Small and Medium Enterprises, and Scientific Research Centres. 13 Early Stage Researchers (ESRs or Fellows) were selected for intensive scientific training in marine and underwater robotics, and it is expected that their training will transform them into desperately needed specialists in underwater technology and qualify them for successful careers in academia and industry. Each fellow was encouraged to enroll into a Doctoral program, and several fellows have completed such programs and received Doctoral degrees before the end of the project.

The scientific research program of Robocademy is divided into three principal action lines:

\begin{itemize}
	\item \textbf{Autonomy}. This covers the ability of a Robot to make the "right" decisions under uncertainty and to be aware of its internal state, with the intent of minimizing human attention and interaction. Four fellows were enrolled in this line, and they developed techniques to automatically detect faults in sensors and actuators, and improved coordination algorithms for multiple robots in order to cooperate to achieve a goal.
	\item \textbf{Disturbance Rejection}. This line deals with the problem of developing sensors and techniques to detect disturbances, and actuators that can be used to compensate and restore normal behaviour. Four fellows worked in this line, developing robust control techniques for underwater manipulators, devising ways for long term navigation under ice with minimum power usage, using machine learning to adaptively learn control model parameters, and developing fish-inspired sensors for flow estimation.
	\item \textbf{Perception}. This final line advances perception techniques for underwater environments. Five fellows developed this program, including neural network techniques for marine debris detection in sonar images, perception using dolphin-inspired acoustic sensors, optical mapping for 3D reconstruction of underwater scenery, and develop 3D reconstruction techniques from Forward-Looking sonar imagery.
\end{itemize}

The fellows have been given a certain amount of freedom to develop their research interests, subject to coordination and approval of their scientific supervisors and the Robocademy scientific board. This has led to some deviation of the research topics, but in general this has resulted in positive contributions, showing how the state of the art has shifted from the initial proposal until the present.

We now present the individual research results for each fellow, categorized by action line.

\section{Action Line Autonomy}

	\subsection{System Monitoring for Persistent Long-Term Autonomy} 
\textbf{Fellow}: Mariela De Lucas Alvarez.\\
\textbf{Host Institution}: Heriot-Watt University, Edinburgh, United Kingdom.\\

System monitoring is a vital component for maintaining the health and integrity of any system. In the field of marine robotics, it plays a vital role in providing long-term autonomy. Autonomous Underwater Vehicles (AUV) require intelligent tools for safe and efficient mission execution. There is an implicit accessibility gap between the AUV and operator during mission execution that requires rules and procedures to be defined prior to deployment. Uncertainties of the environment and unforeseen circumstances can often demand the adaptation or termination of said mission. However, the vehicle cannot adjust its planning or trigger adaptive behaviors to abort or correct abnormalities without accurate identification of the current context. Hence, time and resources of an AUV mission can be more efficiently managed by having precise on-line context information. The line of this work involves the developing of monitoring components that provide self-assessment of the vehicle and status of the mission.

While monitoring tasks can have different objectives to achieve, the line of this work focuses in Fault Detection and Diagnosis (FDD) and Behavior Classification, both approached from a machine learning perspective. Our application of FDD for System Monitoring comprises in implementing methods that learn intrinsic and extrinsic occurrences to the vehicle, i.e a faulty components or mission progression discrepancies caused by environmental elements affecting the performance of the vehicle. 

In conjunction with FDD, Behavior Classification also aids in providing extrinsic context by learning mission related tasks in order to provide and self-asses the evolvement of a mission. Mission monitors have to assess if the ongoing trajectory matches a planned one at a determined point during a mission. For this to be possible, the monitor needs a guideline to compare and recognize if the mission is developing according to plan. To achieve these goals, our work involves the implementation of temporal models for sequential analysis which are widely used for their capabilities of modeling behavior over time. 

The fault detection problem was framed as supervised learning classification. Categorized streams of data from various sensors were used to train a models to classify incoming sequences as types of faults or normal functioning routines. The FDD framework is a Hidden Markov Model with a Gaussian Mixture Model distribution (GMM-HMM) and was used for state monitoring and diagnosis of an AUV thruster emulating system. The system has embedded capabilities for inducing faults and malfunctions and was used to generate a dataset of simulated thruster faults using temperature, voltage and current readings [\cite{delucas2016oceans}]. 

\begin{figure}
	\centering
	\subfloat[Vertical lawnmower and three spiral trajectories.]{ \label{vertical_trj}
		\includegraphics[width=.45\textwidth]{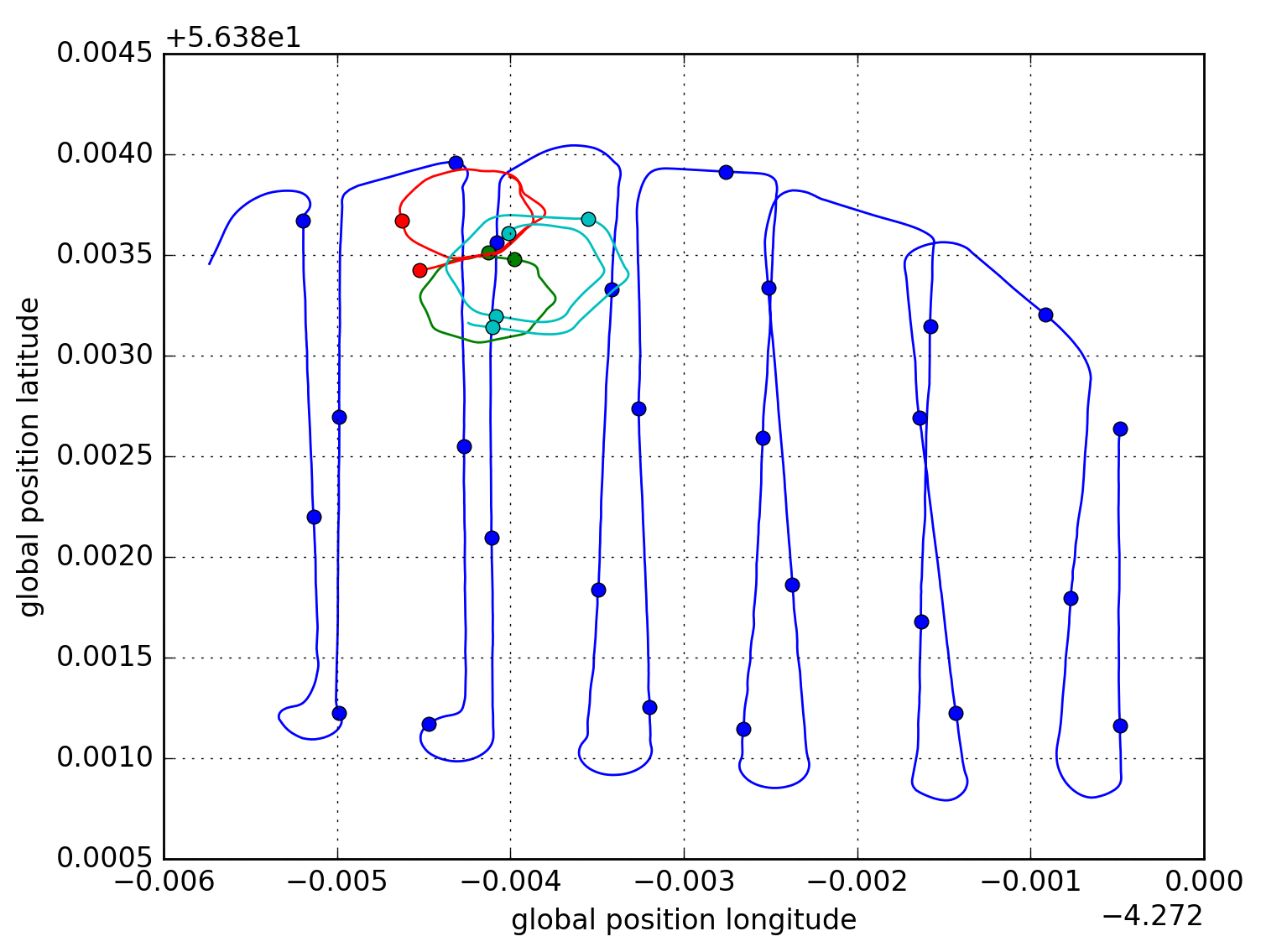}
	}
	\vspace{-0.01cm}
	\subfloat[Diagonal lawnmower and two spiral trajectories.]{ \label{diag_trj}
		\includegraphics[width=.45\textwidth]{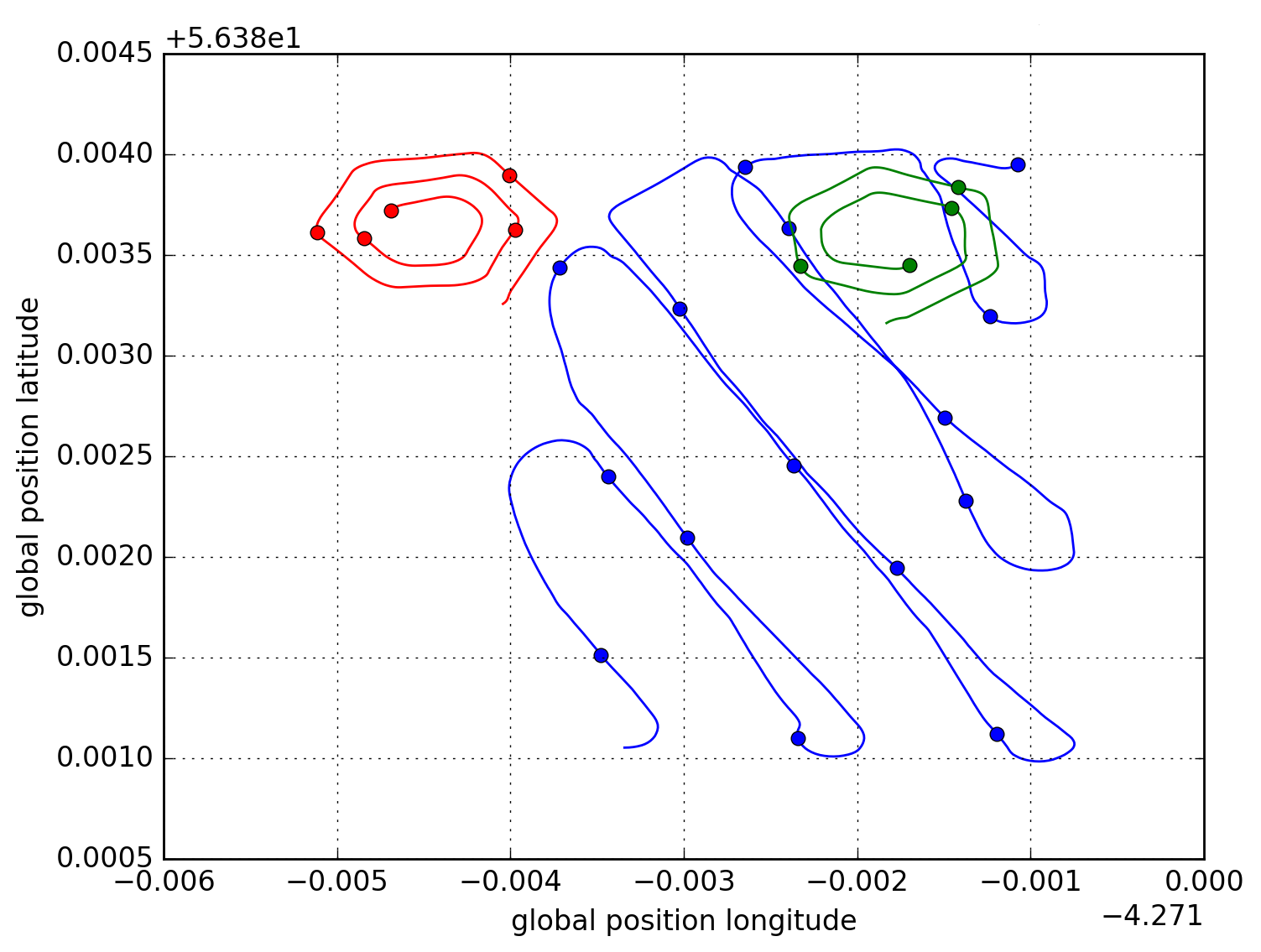}
	}
	\caption{Example of two missions and their extracted trajectory classes. The \textit{Lat} and \textit{Lon} measurements were used, and each sequence is noted with a marker 'o'.}
	\label{fig:PATTERNS}
\end{figure}

Behavior Classification during mission execution was equally framed as a supervised learning problem. The frameworks used were Long Short-Term Memory (LSTM) networks. The data was categorically learned by survey type as part of a known navigation path. The incoming sequences were classified either as lawnmower or spiral inspection trajectories. The two classes of patterns are shown in Figure \ref{fig:PATTERNS}. The data used came from real missions executed by two different AUVs, IVER and REMUS. In this instance, the aim was to classify navigational trajectory patterns in a mine countermeasure (MCM) setting [\cite{delucas2017mlsp}]. The performance for classification of the previously used GMM-HMM and other baseline machine learning models were surpassed when evaluating three different implementations of LSTMs which included hyper-parameter optimization and regularization. 

	\subsection{Planning to Optimize Object Recognition}
\textbf{Fellow}: Veronika Yordanova.\\
\textbf{Host Institution}: Atlas Electronics, Bremen, Germany.\\

Autonomous underwater vehicle (AUV) technology has reached a sufficient maturity level to be considered a suitable alternative to conventional mine countermeasures (MCM) \cite{ferri2017cooperative}. Advantages of using a network of autonomous underwater vehicles include potential time and cost efficiency, no personnel in the minefield, and better data collection. \par 

A major limitation for underwater robotic networks is the poor communication channel \cite{chitre2008underwater}. Currently, acoustics provides the only means to send messages beyond a few metres in shallow water, however the bandwidth and data rate are low, and there are disturbances, such as multipath and variable channel delays, making the communication non-reliable. \par 

One solution for using a network of AUVs for MCM is the Synchronous Rendezvous (SR) method --- dynamically scheduling meeting points during the mission so the vehicles can share data and adapt their future actions according to the newly acquired information \cite{yordanova2015synchronous}. Bringing the vehicles together provides a robust way of exchanging messages, as well as means for regular system monitoring by an operator. \par 

\begin{figure}
        \centering
        \includegraphics[scale=0.45]{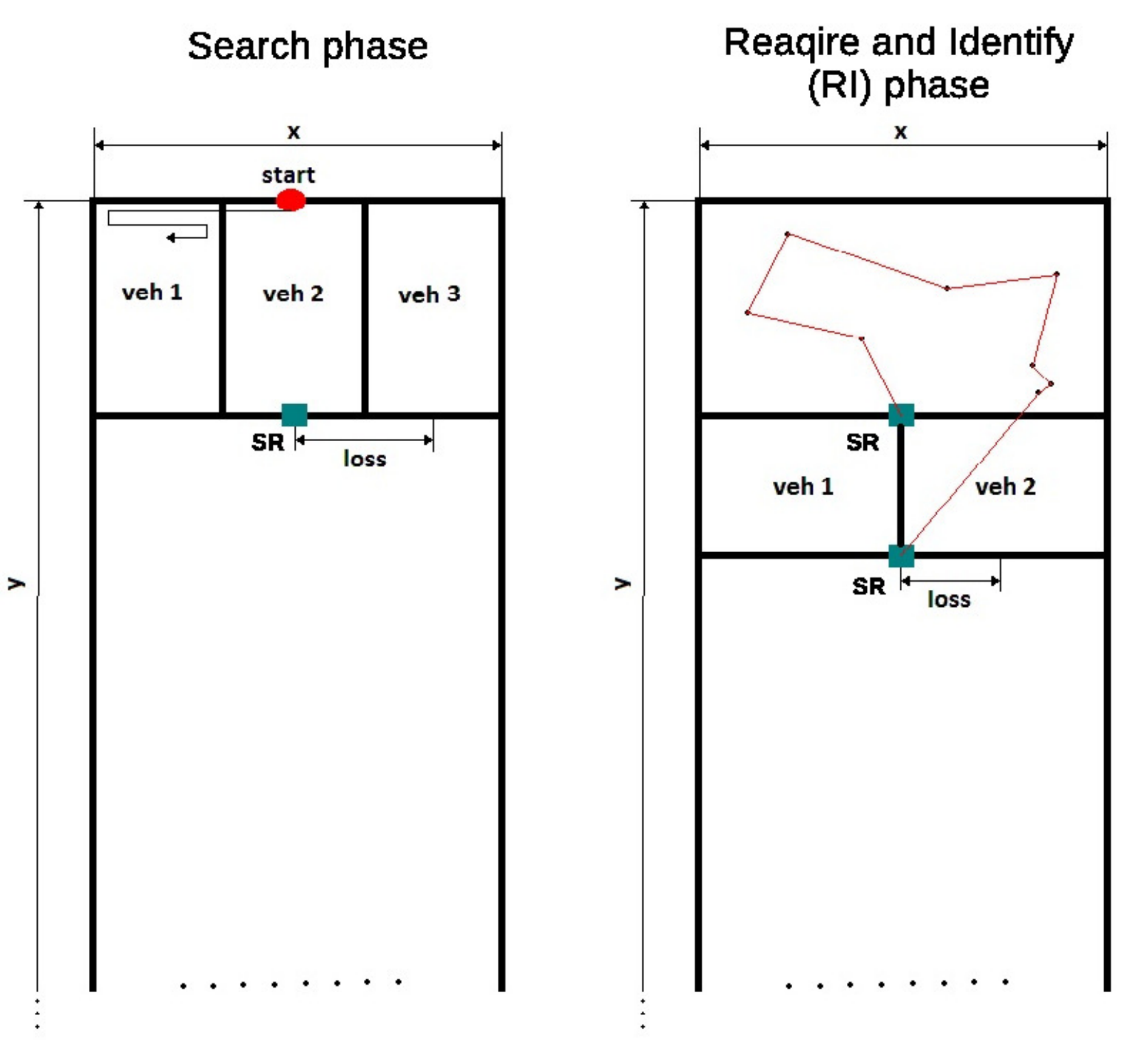}
        \caption [{Synchronous Rendezvous (SR) approach: Left - search phase, Right - reacquiring phase. The area that requires clearing is a rectangle with dimensions $x$ and $y$. The `start` point defines the launching site of the AUVs. On the left graph, three vehicles are moving in a lawnmower pattern. Each node has a separate search area. The vehicles meet at the SR to share status and data gathered during the search phase. On the right, is an example of decision making and retasking of the vehicles. At the SR, they share locations of potential targets that require further inspection.}]{Synchronous Rendezvous (SR) approach: Left - search phase, Right - reacquiring phase. The area that requires clearing is a rectangle with dimensions $x$ and $y$. The `start` point defines the launching site of the AUVs. On the left graph, three vehicles are moving in a lawnmower pattern. Each node has a separate search area. The vehicles meet at the SR to share status and data gathered during the search phase. On the right, is an example of decision making and retasking of the vehicles. At the SR, they share locations of potential targets that require further inspection.}
        \label{search_area_grapth}

    \end{figure}

Figure \ref{search_area_grapth} gives an example of the mission setting and SR method operation. It is aimed at representing a typical MCM mission - demining a strip of seabed near the shore. The `start` point defines the launching site of the AUVs. On the left graph in Figure \ref{search_area_grapth}, the search phase is depicted. Three vehicles are moving in a `lawnmower` pattern, with each having a separate search area. The AUVs meet at the SR to share status and data gathered during the search phase. The graph on the right shows an example of decision making and retasking of the vehicles. At the SR, they have shared locations of potential targets that require further inspection. Based on this information, the vehicles decide to allocate one node to revisit the known targets and two nodes to continue searching new area --- `veh 1` and `veh 2` have a new search pattern and `veh 3` is assigned to revisit objects. The vehicle that revisits the detections follows a greedy shortest path. The time required for this task defines the time of the next SR. The search vehicles adapt their search paths based on their new number, now two instead of three, as well as the new time window --- the SR is adaptive to the workload of revisiting all known objects of interest. This is the advantage of the Synchronous Rendezvous (SR) method --- the robotic network can utilise its resources optimally. The vehicles dynamically adapt their tasks based on the new knowledge developed during the search phase. 

The gains and losses of the SR approach are evaluated against a benchmark scenario of vehicles having their tasks fixed. The numerical simulation results show the advantage of the SR method in handling emerging workload by adaptively retasking vehicles \cite{yordanova2016rendezvous}. \par

The SR method is then further extended into a non-myopic setting, where the vehicles can make a decision taking into account how the future goals will change, given the available resource and estimation of expected workload. Simulation results show that the SR setting provides a way to tackle the high computational complexity load, common  for non-myopic solutions \cite{yordanova2017rendezvous}.\par 
 
Validation of the SR method is based on trial data and experiments performed using a robotics framework, MOOS-IvP. \par 

This work develops and evaluates the Synchronous Rendezvous method, a mission planning approach for underwater robotic cooperation in communication and resource constraint environment \cite{yordanova2018phd}.

\section{Action Line Disturbance Rejection}

	\subsection{Robust Control Strategies for Single and Multiple Autonomous Underwater Vehicles}
\textbf{Fellow}: Shahab Heshmati-Alamdari.\\
\textbf{Host Institution}: National Technical University of Athens, Athens, Greece.\\

In this work \cite{Heshmati-Alamdari1315117}, we address the problem of robust control for underwater robotic vehicles under resource constraints inspired by practical applications in the field of marine robotics. By the term ``resource constraints" we refer to systems with constraints on communication, sensing and energy resources. Within this context, the ultimate objective of this work lies in the development and implementation of efficient model and control strategies for autonomous single and multiple underwater robotic systems considering significant issues such as: external disturbances, limited power resources, strict communication constraints along with underwater sensing and localization issues.

Specifically, we focused on cooperative and interaction control for single and multiple Underwater Vehicle Manipulator Systems (UVMSs) considering the aforementioned issues and limitations, a topic of utmost challenging area of marine robotics. More precisely, the contributions of this work lie in the scope of four topics: \emph{i) Motion Control}, \emph{ii) Visual servoing}, \emph{iii) Interactioncontrol} and \emph {iv)Cooperative Transportation}.

\begin{figure}[!htbp]
	\begin{center}
		\includegraphics[width=3.3in]{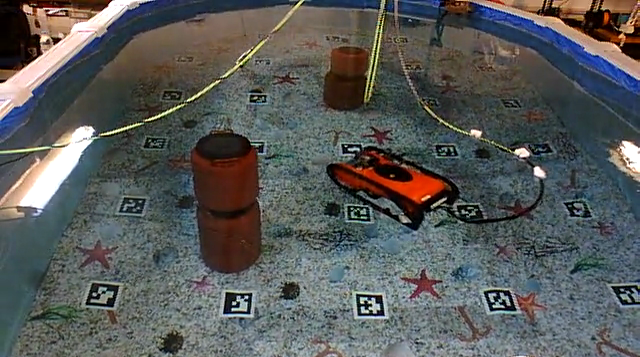}
	\end{center}
	\caption{Experimental setup and problem formulation: the purpose of the controller is to guide the vehicle towards desired way points inside a constrained workspace including sparse obstacles.  }\label{new_fig1}
\end{figure}

\noindent \textbf{{Motion control:}}\vspace{1mm}\\
we formulated in a generic way the problem of Autonomous Underwater Vehicle (AUV) motion operating in a constrained environment including obstacles. Specifically, during the control design, we considered strictly significant limitations such as: energy consumption issues, robustness with respect to external disturbances, dynamic parameter uncertainties, thruster saturations, obstacles, workspace boundary, predefined upper bound of the vehicle velocity\footnote{requirements for various underwater tasks such as seabed inspection, mosaicking etc \cite{Bechlioulis2017429}.}. The obstacle avoidance has been designed based on the system`s sensing ranges. This has allowed the AUV to compute or update its path in real-time based on the detected obstacles as the AUV moves through the workspace. Moreover, the controller has been designed in a way that the vehicle exploits the ocean currents, which results in reduced energy consumption by the thrusters and consequently increases significantly the autonomy of the system. The closed-loop system has been guaranteed stability and convergence properties. The performance of the proposed control strategy was experimentally verified using a 4 Degrees of Freedom (DoF) underwater robotic vehicle inside a constrained test tank in the Control System lab of the Mechanical Engineers Dept. of the National Technical University of Athens (Fig.~\ref{new_fig1}). A detailed description of this work is reported in \cite{heshmatiTCST,heshmatirobusticra2018}. This work was next extended in \cite{Heshmati_CDC2019} by proposing a robust trajectory tracking control scheme for underactuated AUVs operating inside a partially known and dynamic environment where the knowledge of the operating workspace is constantly updated via the vehicle’s on–board sensors.

\noindent \textbf{{Visual servoing:}}\vspace{1mm}\\
During this part (i.e., for cases that the robot is close to the point of interest), we consider visual feedback (i.e., imaging sonar or usual camera) as an appropriate feedback for designing of efficient controllers. This is motivated by the delay and inaccuracy of the measurements provided by common underwater localization systems e.g., Doppler Velocity Log (DVL) or Ultra Short Baseline (USBL).

More specifically, we formulated a number of novel visual servoing control strategies (i.e., Image based and Position based) in order to reach the robot (robot’s end effector) close to the point of interest. Within this framework, we have considered significant issues in visual servoing control such as: camera Field of View (FoV) constraint, robustness with respect to Camera Calibration uncertainties and the resolution of visual tracking algorithm. Specifically, first, based on Nonlinear Model Predictive Control (NMPC)\footnote{For more details on NMPC, please refer to \cite{allgower2004,Eqtami2013638,Eqtami20137384,Logothetis2018,Nikou2017707,Nikou2018990} and the paper cited and papers quoted therein.}, we designed a Position Based Visual Servoing (PBVS) scheme for case when the relative position between the robot and the point of interest is available online (e.g., by detection of a known marker on the object).  A detailed description of this work is reported in \cite{Heshmati-Alamdari20143826}.

Then we extended this work to an Image Based Visual Servoing (IBVS) \cite{Heshmati-Alamdari20144469} scheme for case that the relative position is not available online. Both of the aforementioned schemes are combined with a self-triggering mechanism that decides when the Visual Tracking Algorithm needs to be triggered and new control signals must be calculated. Thus, the proposed schemes result in reduction of the CPU computational effort, energy consumption and increases the autonomy of the system. A detailed description of this work is reported in \cite{Heshmati-Alamdari20155492}.

 Furthermore, for case when neither the Camera Calibration Matrix of the supposed vision system is available in priory, a model free IBVS control strategy is proposed which that imposes prescribed transient and steady state response on the image feature coordinate errors and satisfies the visibility constraints that inherently arise owing to the camera’s limited field of view, despite the inevitable calibration and depth measurement errors. The computational complexity of the proposed schemes proves significantly low. It is actually a static scheme involving very few and simple calculations to output the control signal, which enables easily its implementation on fast embedded control platforms. A detailed description of this work is reported in \cite{PPIBVS_TRO,Heshmati-Alamdari20142721}. \vspace{4mm}

\noindent \textbf{{Interaction Control}}\\

It is well known that underwater tasks are very challenging owing mainly to external disturbances (i.e., sea currents), the lack of appropriate and adequately accurate sensing/localization  and the unknown (or partially known) constrained environment (e.g., offshore industry, oil/gas facilities) \cite{Hurtos20141978}. The aforementioned difficulties make the control of underwater manipulator systems a challenging problem that has already gained significant scientific attention within the marine robotic community during the last years.

\begin{figure}[!htb]
	\centering
	\setlength{\fboxsep}{0pt}%
	\setlength{\fboxrule}{2pt}%
	\includegraphics[scale=0.23]{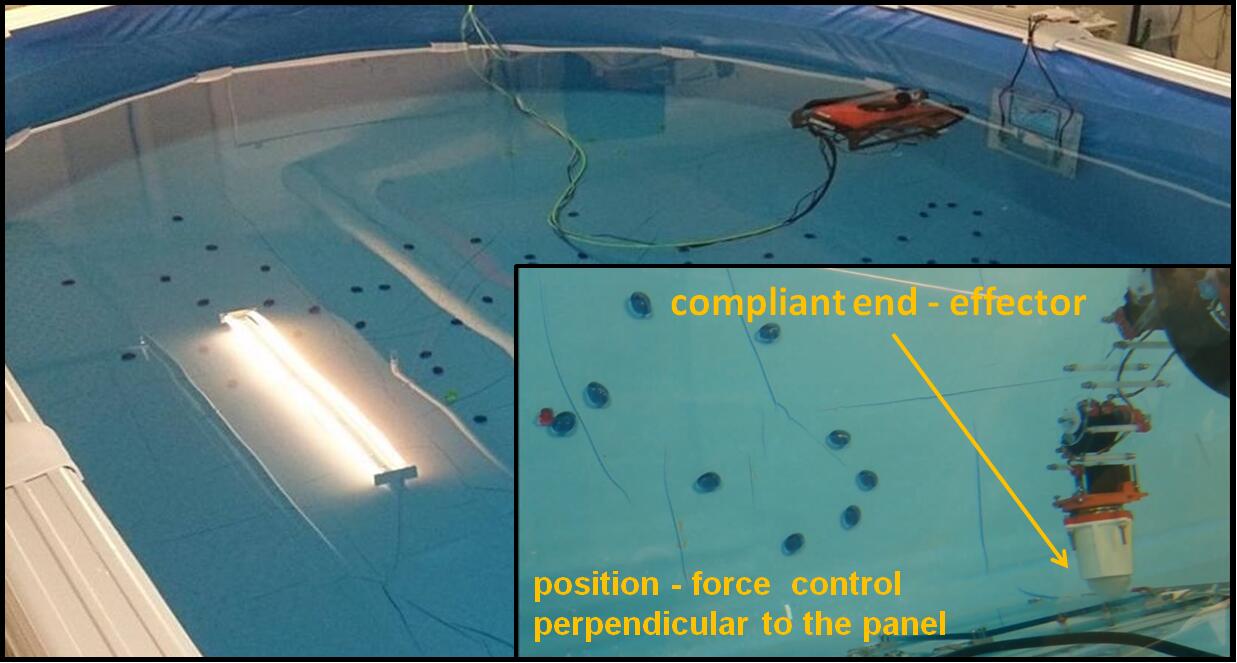}
	\caption{The \textit{NTUA, Control Systems Lab} test tank and Position--Force Control Scenario }
	\label{fig:exp_env}
\end{figure}

Generally, in underwater robotic interaction tasks various issues regarding the uncertainties and complexity of the robot dynamic model, the external disturbances (e.g., sea currents), the steady state performance as well as the overshooting/undershooting of the interaction force error, should be addressed during the control design. Motivated by the aforementioned considerations, regarding to the interaction part, we present a robust interaction control scheme for a UVMS in contact with the environment, with great applications in underwater robotics (e.g. sampling of the sea organisms, underwater welding, object handling). The proposed control scheme does not required any a priori knowledge of the UVMS dynamical parameters or the stiffness model. It guarantees a predefined behavior in terms of desired overshoot, transient and steady state response and it is robust with respect to external disturbances and measurement noises. Moreover, the proposed control framework guarantees: (i) certain predefined minimum speed of response, maximum steady state error as well as overshoot/undershoot concerning the force/position tracking errors, (ii) contact maintenance and (iii) bounded closed loop signals. Additionally, the achieved transient and steady state performance is solely determined by certain designer-specified performance functions/parameters and is fully decoupled from the control gain selection and the initial conditions. Simulation and experimental studies clarify the proposed method and verify its efficiency (See Fig \ref{fig:exp_env}).  A detailed description of this work is reported in \cite{Heshmati-ARC,Heshmati-alamdari201711197}.

\noindent \textbf{{Cooperative Transportation}}\vspace{2mm}\\
Most of the underwater manipulation tasks can be carried
out more efficiently, if multiple UVMSs are cooperatively involved. Underwater tasks are very demanding, with the most significant challenge being imposed by the strict communication constraints. In general, the communication of multi-robot systems can be classified in two major categories, namely explicit (e.g., conveying information such as sensory data directly to other robots) and implicit (e.g., the interaction forces between the object and the robot). Even though the inter-robot communication is of utmost importance during cooperative manipulation tasks, employing explicit communication in underwater environment may result in severe performance problems owing to the limited bandwidth and update rate of underwater acoustic devices. Moreover, as the number of cooperating robots increases, communication protocols require complex design to deal with the crowed bandwidth. Cooperative manipulation has been well studied in the literature, especially the centralized schemes. Despite its efficiency, centralized control is less robust, since all units rely on a central system. On the other hand, the decentralized cooperative manipulation schemes depend usually on either explicit communication interchange among the robots (e.g., online transmission of the desired trajectory or off–line knowledge of the objects’ trajectory). Therefore, the design of decentralized cooperative manipulation algorithms for underwater tasks employing implicit and lean explicit communication becomes apparent.

\begin{figure}[!htb]
	\centering
	\setlength{\fboxsep}{0pt}%
	\setlength{\fboxrule}{2pt}%
	\includegraphics[scale=0.23]{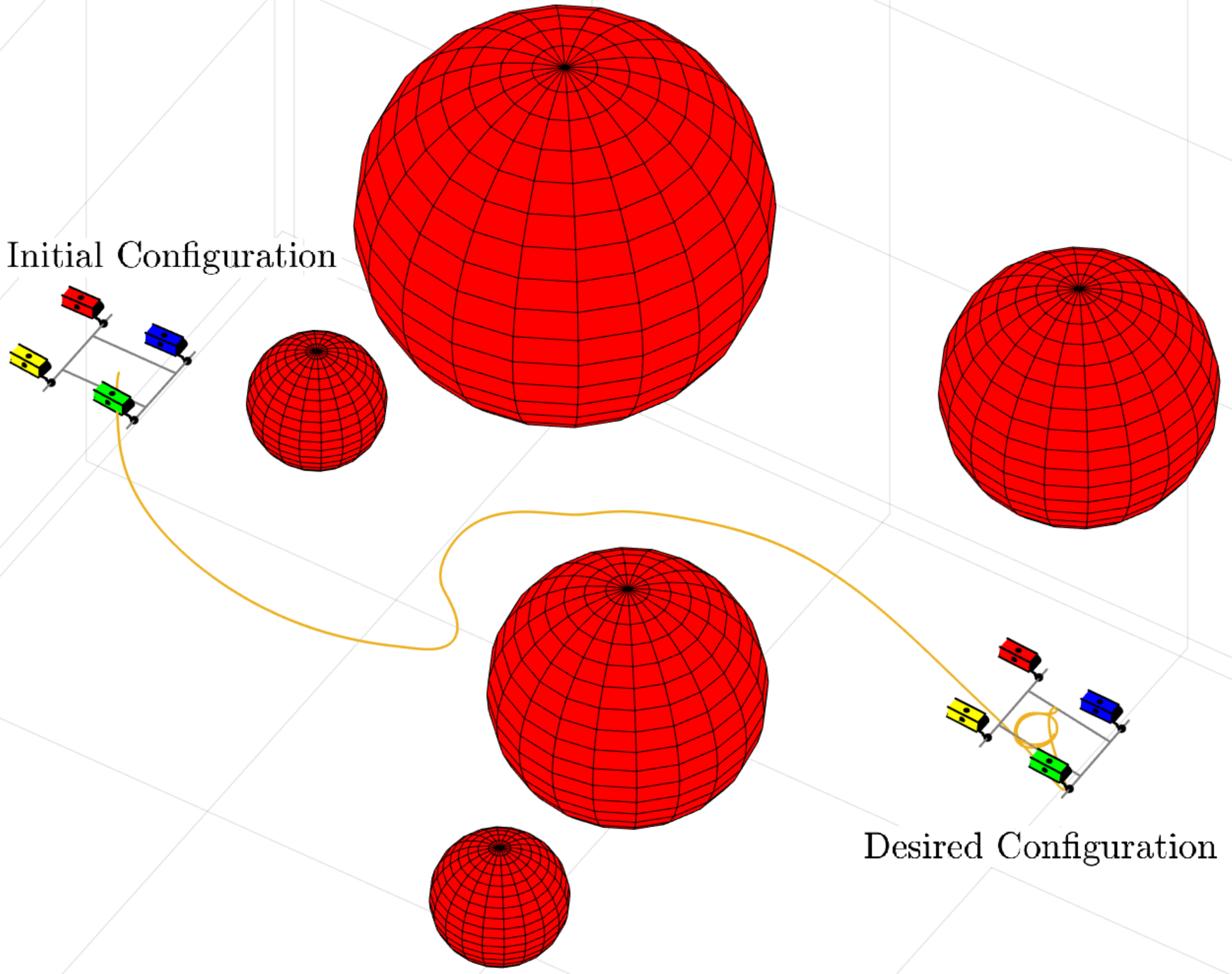}
	\caption{Cooperative object transportation for a team of UVMSs in a constrained workspace. }
	\label{fig:workspace}
\end{figure}

Based on the aforementioned issues, first  we addressed the problem of cooperative object transportation for multiple Underwater Vehicle Manipulator Systems (UVMSs) in a constrained workspace involving static obstacles (See Fig \ref{fig:workspace}), with the coordination relying solely on implicit communication arising from the physical interaction of the robots with the commonly grasped object. We propose a novel distributed leader-follower architecture, where the leading UVMS, which has knowledge of the object’s desired trajectory, tries to achieve the desired tracking behavior via an impedance control law, navigating in this way, the overall formation towards the goal configuration while avoiding collisions with the obstacles. On the other hand, the following UVMSs estimate locally the object’s desired trajectory via a novel prescribed performance estimation law and implement a similar impedance control law achieving in this way tracking of the desired trajectory despite the uncertainty and external disturbance in the object and the UVMS dynamics respectively. The feedback relies on each UVMS’s force/torque measurements and no explicit data is exchanged online among the robots, thus reducing the required communication bandwidth and increasing robustness. Moreover, the control scheme adopts load sharing among the UVMSs according to their specific payload capabilities. A detailed description of this work is reported in \cite{Heshmati_Oceanic2018, Heshmati_AUV2018}.

\begin{figure}[!htb]
	\centering
	\subfloat[Simulation environment]{ \label{fig:shahab_ex_a}
		\includegraphics[trim=0 0 0 0,clip,width=0.45\textwidth]{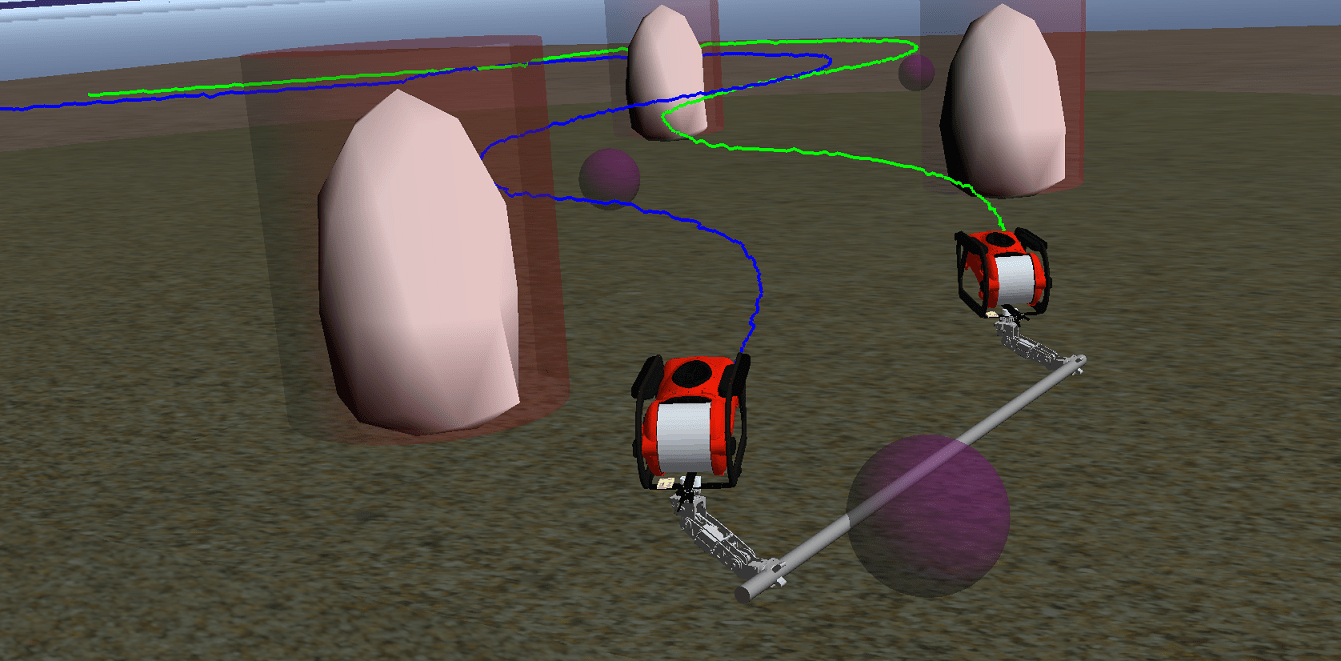}
	}
	\vspace{-0.01cm}
	
	\subfloat[Real experiment]{ \label{fig:shahab_ex_b}
		\includegraphics[trim=0 30 0 30,clip,width=0.45\textwidth]{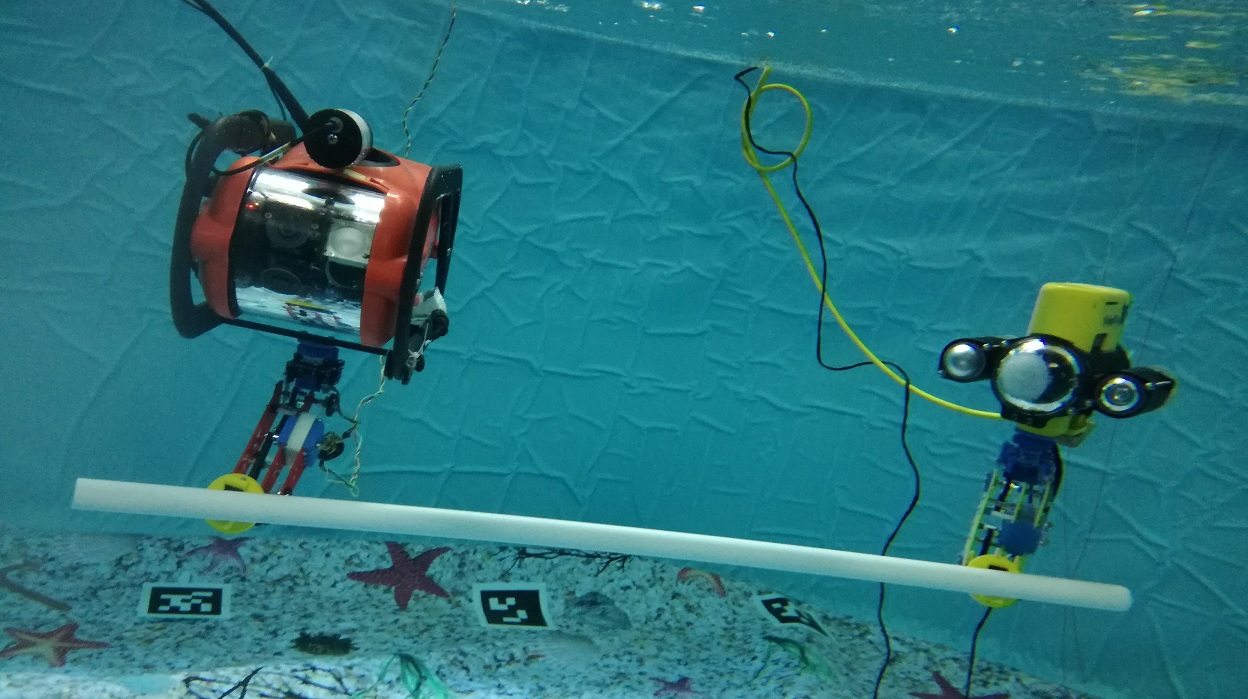}
	}
	\caption{Cooperatively object transportation using two UVMSs inside a constrained workspace including obstacles.}
	\label{fig:shahab_ex}
\end{figure}

Moreover we extended the aforementioned work proposed a Nonlinear Model Predictive Control (NMPC) approach for a team of UVMSs in order to transport an object while avoiding significant constraints and limitations such as: kinematic and representation singularities, obstacles within the workspace, joint limits and control input saturation. More precisely, by exploiting the coupled dynamics between the robots and the object, and using certain load sharing coefficients, we designed a distributed NMPC for each UVMS in order to cooperatively transport the object within the workspace’s feasible region. Moreover, the control scheme has been designed in order to adopt load sharing among the UVMSs according to their specific payload capabilities. Additionally, the feedback in the proposed scheme relies on each UVMS’s locally measurements and no explicit data is exchanged online among the robots, thus reducing the required communication bandwidth. Finally, real-time simulation (See Fig. \ref{fig:shahab_ex_a}) results conducted in UwSim dynamic simulator running in ROS environment as well as real time experiments (See Fig. \ref{fig:shahab_ex_b}) performed within a test tank verify the effectiveness of the theoretical finding. A detailed description of this work is reported in \cite{Heshmati_2018_MPC_coop,Heshmati_ICRA2019}.

	\subsection{Self-Adapting Motion Models of Underwater Vehicles}
\textbf{Fellow}: Bilal Wehbe.\\
\textbf{Host Institution}: German Research Centre for Artificial Intelligence, Bremen, Germany.\\

Autonomous underwater vehicles (AUVs) are robotic systems designed as an
alternative to manned submerged vehicles, to carry out specific and well-defined
underwater tasks. The application of AUVs range from scientific expedition
missions and data collection to commercial and industrial purposes. As part of
their development process, modeling the hydrodynamic behaviour of these vehicle
is a crucial step for analyzing their stability, implementations of model-based
control schemes, developing high fidelity simulators, and aiding their
navigation systems. This model is developed by analyzing the dynamics of the
fluid surrounding the body, which involves estimating the motion states as
function of their control inputs and other environmental aspects.

Moreover, AUVs can generally succeed in performing missions when operating in
known and stable environments and when their navigation sensors are in good
condition, but when the environmental situations change or in cases of sensors
malfunction and drop-outs, these robots can get easily confused or lost. This
problem get more severe when AUVs are designed to undertake long-term missions
and operate for prolonged periods of time.
Environmentally induced changes such as temperature, viscosity, or density
fluctuations of the water and bio-fouling, or technical changes such as
variation of vehicle's payload, adding or removing of external components, body
wear and damage, actuator malfunctions or failure, are all factors that affect
the model directly and can lead to a change in the robot's expected behavior.
Thus in practice, even a good working static model would fail to predict correctly
the states of the vehicle when its actual dynamics change.

In this work, we tackle these problems by developing a framework that first provides
an efficient data-driven method that learns the dynamic model of an AUV directly
from its onboard navigation sensors. Second, the framework provides a method to
learn and adapt the model in real-time while the robot is operating, and
therefore correct the mismatches between the model predictions and the data when
the dynamics of the AUV changes.  Third, it develops an intelligent observer
that monitors the changes of the vehicle's dynamics and thus be able to classify
the anomalies occurring during robot operation.

\begin{figure}
	\centering
	\subfloat[Leng]{ \label{fig:leng}
		\includegraphics[trim=0 0 0 0,clip,width=0.45\textwidth]{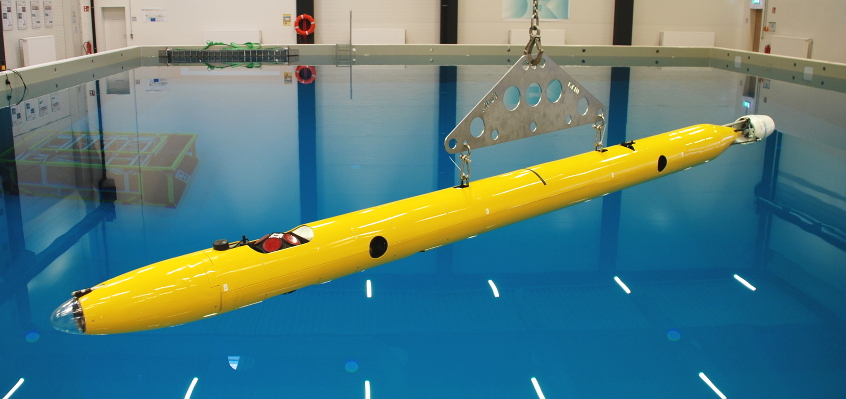}
	}
	\vspace{-0.01cm}
	
	\subfloat[Dagon]{ \label{fig:dagon}
		\includegraphics[trim=0 30 0 30,clip,width=0.45\textwidth]{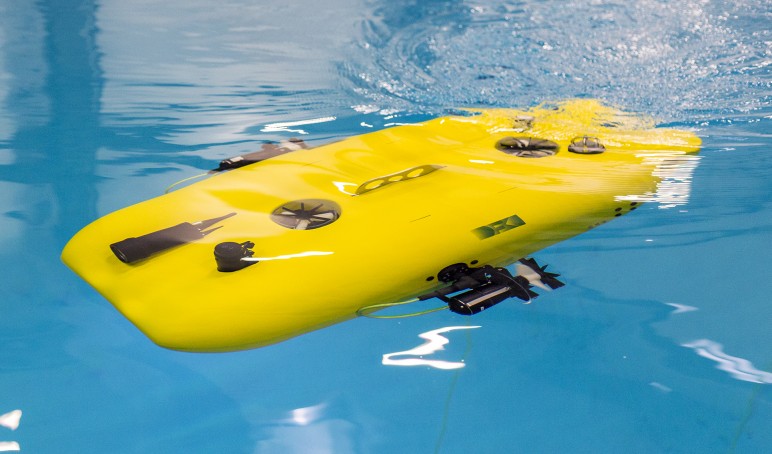}
	}
	\caption{AUVs developed at DFKI-RIC labs used in experimental setup.}
	\label{fig:lenghull}
\end{figure}

For realizing this framework, several experiments were conducted using two AUVs
Leng (Fig.~\ref{fig:leng}) and Dagon (Fig.~\ref{fig:dagon}) developed at the
labs of DFKI - RIC in Bremen, Germany. Data collected from this experimental
setup was used to estimate the motion models of these vehicles. The performance
of several machine learning regression methods were tested against the classical
physics-based approach to model the robots' hydrodynamic models. We report a
detailed description of this work in \cite{wehbe-icra17}, where we conclude that
nonlinear regression methods show better capabilities than physics-based methods
to model the hydrodynamical properties of underwater vehicles.
In \cite{wehbe-oceans17} this work was extended to include multi-DOF coupled
models. A data driven approach which employs a machine learning technique known
as Support Vector Regression (SVR), to identify the coupled dynamical model of
an AUV. The approach was based on a variant of a radial-basis-function kernel
is used in combination with the SVR which accounts for the different
complexities of each of the contributing input features of the model.

In \cite{Wehbe2017b} we developed an online technique which employs incremental
SVR to learn the damping term of an underwater vehicle motion model, subject to
dynamical changes in the vehicle’s body. We a new sample-efficient methodology 
was introduced which accounts for adding new training samples, removing old 
samples, and outlier rejection. The proposed method is tested in a real-world
experimental scenario to account for the model’s dynamical changes due to a
change in the vehicle’s geometrical shape.

	\subsection{Terrain-Aided Navigation and Ultra-Endurance AUVs for Crossing the Arctic Ocean}
\textbf{Fellow}: Georgios Salavasidis.\\
\textbf{Host Institution}: National Oceanography Centre, Southampton, United Kingdom.\\

\begin{figure}[h!]
	\centering
	\includegraphics[width=0.5\textwidth]{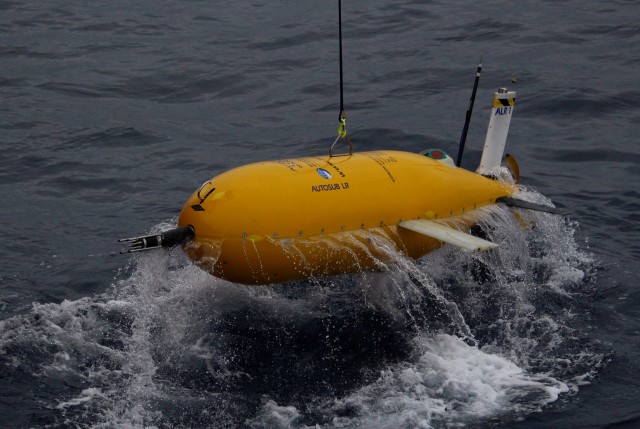}
	\caption{Autosub Long Range, a long-endurance and 6000m deep rated AUV. Recovery after a deep water and long range mission in the Southern Ocean.}
	\label{fig:ALR}
\end{figure}

\begin{figure}[h!]
	\centering
	\includegraphics[width=0.5\textwidth]{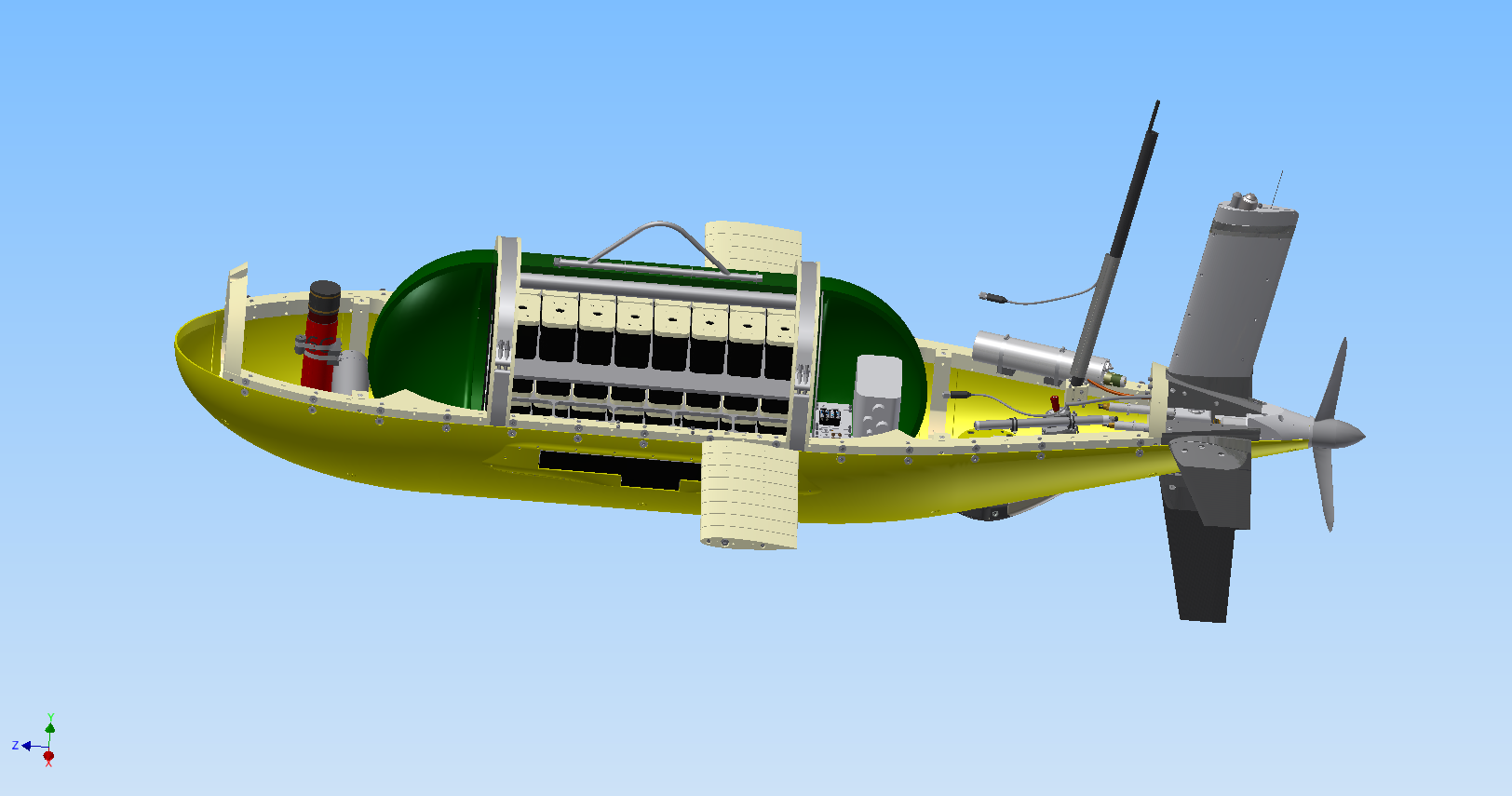}
	\caption{Autosub Long Range 1500, an ultra-endurance 1500m depth rated autonomous underwater vehicle.}
	\label{fig:ALR-1500}
\end{figure}

With the recent developments of ultra-endurance AUVs capable of operating for several months, such as the Autosub Long Range (ALR)~(Fig.~\ref{fig:ALR}) and the Autosub Long Range 1500 (ALR-1500)~(Fig.~\ref{fig:ALR-1500}) reported in \cite{furlong2012autosub} and~\cite{Autosub1500}, a world of new AUV applications is opened up, including persistent monitoring and data collection in some of the remotest and deep areas on Earth. One example mission for the such long-range platforms is the crossing of the Arctic Ocean investigated in~\cite{IcraGeorgios}. 

However, the underwater localisation problem becomes increasingly challenging as the operation time increases. Although inertial navigation provides sufficient short term localisation accuracy, the performance degrades over time due to sensor inaccuracies. The most common method to bound the inertial navigation error growth is using acoustic-based aiding systems as in~\cite{Munafo} and~\cite{salavasidis2016co}. Despite the accuracy and robustness of these methods, the reliance on external infrastructure restricts the AUVs to operate within an area of few kilometres due to limitations imposed by physical constraints (signal attenuation). Alternatively, the localisation drift can be bounded by regular surfacing. This option, however, is not available for operations under the 12\% of the world’s oceans covered by ice. To operate effectively in such regions requires an entirely on-board and robust localisation solutions. By relying only on on-board sensors and estimation algorithms, the navigation solution must fulfil two main requirements: a) bounded navigation error, b) conform to energy constraints and conserve on-board power.

One way to face the challenge of unaided navigation and place a bound on the positing error growth is to use Terrain-Aided Navigation (TAN). TAN is a form of geophysical localisation where water depth measurements are matched against a priori bathymetric reference map as in~\cite{salavasidis2016terrain}. In general terms, the idea is to exploit physical features of the seabed morphology in order to estimate the position of the AUV within this map. To undertake this task, statistical estimators are used to generate estimates of the true vehicle position given a sequence of water depth measurements, see~\cite{melo2017survey} for an overview of the current status. In this respect, Bayesian estimation is a widely used framework for vehicle localisation because of its efficiency in fusing multiple sources of information, including both statistics and uncertainty models. In contrast to acoustic-aided methods, TAN falls in the class of completely on-board localisation methods, where there is no need for external aiding infrastructures. This property makes these methods particularly appealing for long range AUVs.

\begin{figure}[htb]
	\centering
	\includegraphics[width=0.5\textwidth]{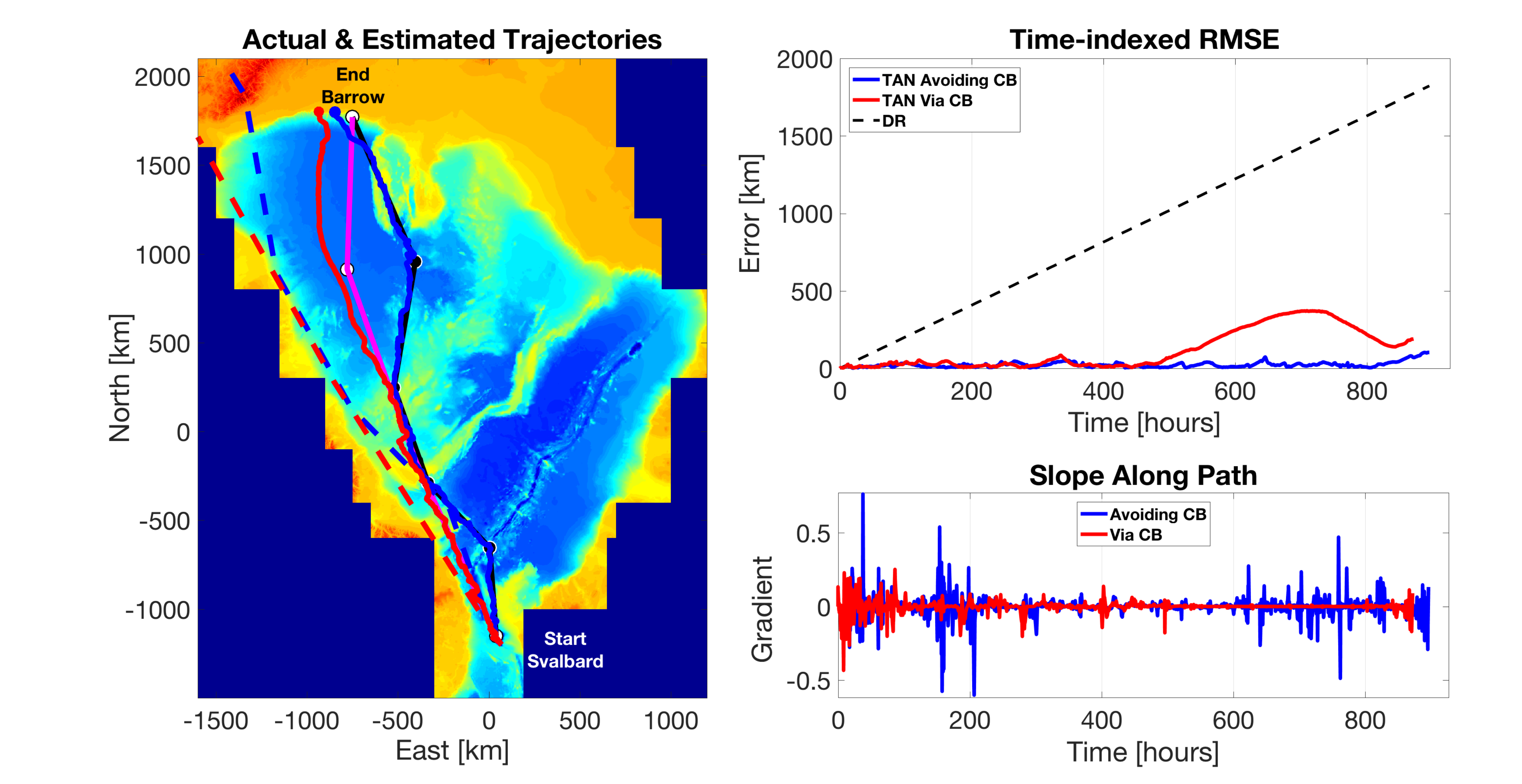}
	\caption{\underline{Left:} Actual and estimated (using TAN and DR) trajectories for the two Arctic crossing scenarios. Blue lines (solid for TAN and dashed for DR) correspond to position estimates while the actual trajectory (black) avoids CB. Red lines (solid for TAN and dashed for DR) correspond to position estimates while the actual path crosses CB (magenta). \underline{Right Top:} RMS error over time for TAN and DR. \underline{Right Bottom:} Bathymetric slope along the two Arctic crossing scenarios.}
	\label{fig: TrajectoryErrorANDGradient}
\end{figure}

In~\cite{IcraGeorgios} a tractable and low complexity particle filter-based TAN algorithm has been developed for ultra-endurance AUVs. This research investigates the potential of the TAN to bound the positioning error while the ALR-1500 attempts to cross the Arctic Ocean, a distance greater than 3500 km and approximately 2 months operation. To satisfy the energy constraints for such long-range missions, the AUV has restrictions in the forward speed, the number and quality of localisation sensors and the processing power for navigation algorithms. The navigation filter estimates only the vehicle horizontal position by fusing readings from the motion sensors and a low frequency single-beam echo-sounder. The algorithm is evaluated in simulation and the effect of various environmental characteristics, sensor limitations and filter-related parameters, affecting the estimation accuracy and robustness is demonstrated. To highlight the importance of terrain variation for the TAN, the crossing of the Arctic Ocean (from Svalbard to the Barrow coast in Alaska) is performed along two morphologically different trajectories, see Fig.~\ref{fig: TrajectoryErrorANDGradient}. The westerly trajectory defines a low informative path via the Canada Basin (CB), which is a deep and relatively flat oceanic basin, while the easterly trajectory avoids the CB.
The simulation is set up taking into account the worst excepted scenarios (in terms on localisation) and the estimation error results show that the TAN is able to address the localisation problem even in the case of limited resources and a very sparse and noisy bathymetric map, while estimates using Dead Reckoning (DR) techniques experience dramatic error growth over time. The terrain morphology analysis shows that the algorithm exhibits significantly higher accuracy and consistency if the AUV is guided over morphologically informative regions.

Towards our ultimate goal of crossing the Arctic Ocean, TAN algorithms are further being tested using real data. In~\cite{FoRGeorgios} we develop and evaluate the accuracy and robustness of a TAN algorithm using field data from three deep (up to 3700 m) and long range (up to 195 km and 77 hours) missions of the ALR in the Southern Ocean. The particle filter-based algorithm is designed and built in a way to accommodate the needs for extended duration underwater operations by providing light and tractable navigation solution. The filter parameters are set up based on environmental characteristics and on-board processing power. To assess the a posteriori TAN performance, TAN position estimates are compared with the real-time estimates (via dead-reckoning) and USBL measurements, which are considered as baseline positions, intermittently taken from a support ship throughout the missions. Despite the challenges, such as low number and quality sensors, motion along bathymetric contour-lines for long-time intervals and occasionally poor quality (or even lack) of water depth measurements, the algorithm maintains convergence and bounded error in all three mission and, in overall, high consistency between 20 Monte Carlo runs. 

Both simulation and operational-data based results illustrate that the TAN has the potential to prolong underwater missions to a range of thousands of kilometres without the need for surfacing and re-initialising the estimation process, given sufficient terrain variability and proper sensor and environmental characterisation. Future research will focus on three key areas. Firstly, the filter robustness will be enhanced by incorporating strategies for divergence detection and filter re-convergence. Secondly, research will focus on developing probabilistic reference map models. Thirdly, attention will be directed towards developing path planning algorithms for directing AUVs towards morphologically informative regions of the Arctic Ocean. 

	\subsection{Flow Sensing}
\textbf{Fellow}: Juan Francisco Fuentes-Perez.\\
\textbf{Host Institution}: Tallinn University of Technology, Tallinn, Estonia.\\

To effectively take decisions and move in underwater environments it is necessary to gather information from it, sense it. The most common sensing modalities in underwater vehicles are vision and sonar. However, both sensing modalities might be of little utility in some environments such as homogeneous or turbid ones. Therefore, the combination of multiple sensing modalities, or to develop new sensing technologies, can be the most effective solutions.

\begin{figure}[b]
	\centerline{\includegraphics{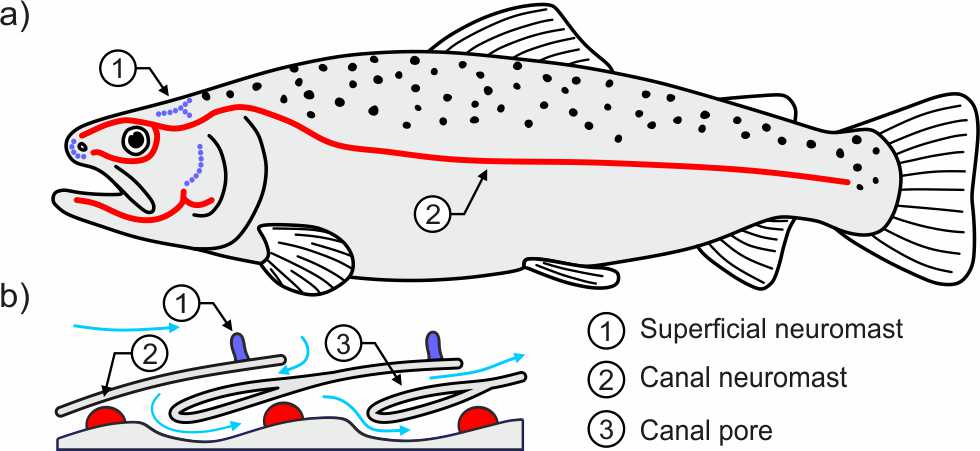}}
	\caption{Lateral line system (\cite{Tuhtan2017}); (a) Distribution of neuromasts; (b) Superficial and canal neuromasts.}
	\label{Fig1_FlowSensing}
	
\end{figure}

To solve these limitations researchers have tried to inspire by fish (\cite{Salumae2015,Jezov2013}). Fish has evolved in water and thus they present unique adaptations to live in this media. One of those adaptations is the lateral line ((Fig.~\ref{Fig1_FlowSensing}). The lateral line is a sensing organ able to obtain data from the flow, which after is encoded in the nervous system of the fish to translate it to useful information, allowing fish to act and react accordingly to what they sense. Therefore, lateral lines are responsible for different fish behaviors, such as rheotaxis, schooling, predator and obstacle avoidance, prey localization, as well as stationary obstacle detection to reduce energy consumption (\cite{Cook2010,Yang2010}). 

\begin{figure}[b]
	\centerline{\includegraphics{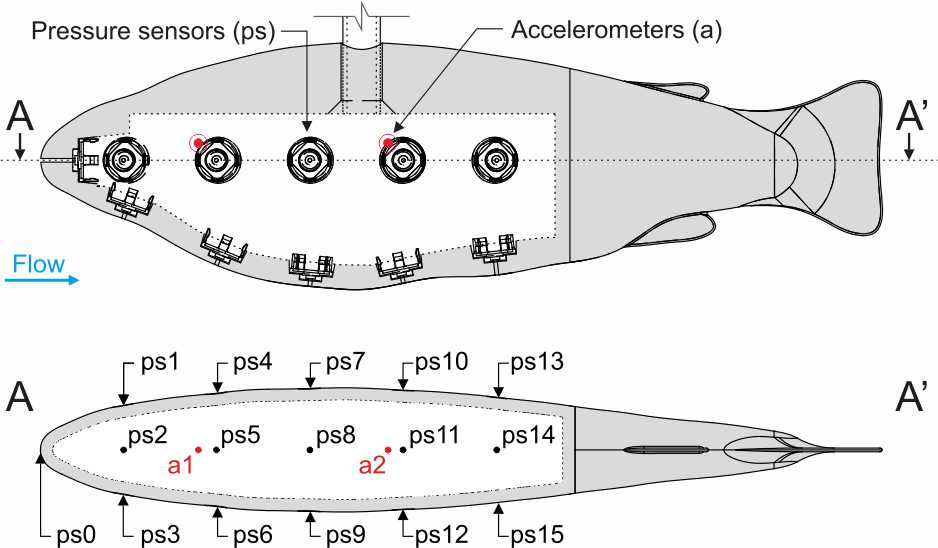}}
	\caption{Distribution of the pressure sensors in the prototype used in the present study. ps0 is the nose sensor, and ps1 to ps15 are the lateral sensors (\cite{Fuentes-Perez2015}).}
	\label{Fig2_FlowSensing}  
\end{figure}

Inspired by these natural organs, researchers have developed artificial lateral lines (ALL). As its organic analog, the ALL is usually designed as a discrete set of sensing units placed over a body where they record the local mechanical changes in water (\cite{Dijkgraaf1963}) ((Fig.~\ref{Fig2_FlowSensing}). Many researching groups had studied this technology and augured incredible opportunities and advances for underwater robotics with the use of it (\cite{Yang2010,Fan2002,Chen2006,Yang2006,Pandya2006}). However, to date, these devices have only demonstrated satisfactory performance under ideal hydrodynamic conditions (\cite{Yang2010, Chen2006,Yang2006,Venturelli2012,Salumae2013}) and, less frequently, their performance has depended on laboratory models that would be difficult to apply in real scenarios (\cite{Fan2002, Chen2006,Yang2007,Abdulsadda2012}). Therefore, it is necessary to advance in their study to achieve a robust device able to cope with the limitations of field conditions, which is able to measure complex hydrodynamic scenarios, and collect data about underwater environments, that after can be used to understand and exploit better the underwater media and control underwater vehicles.

This researching project tries to solve the above-mentioned problems and advance in the knowledge of underwater environments by exploring the limits of pressure sensor based ALL. For this, an incremental development is followed (\cite{Fuentes-Perez2018a}). First, absolute pressure sensor based ALL under complex hydrodynamics situations are studied. Classical methods that make use of pressure for hydrodynamic variable estimation are studied and analyzed and new methods, as well as combined methods, are proposed to adapt their performance to more hostile environments (\cite{Fuentes-Perez2015,Chen2017,Tuhtan2017,Strokina2016}). This analysis demonstrates their utility to solve real-like problems and they utility as an alternative or complement of other sensing modalities (\cite{Fuentes-Perez2017b}), but also their limitations.

\begin{figure*}[t!]
	\centerline{\includegraphics[trim=0 0 0 0,clip,width=1\textwidth]{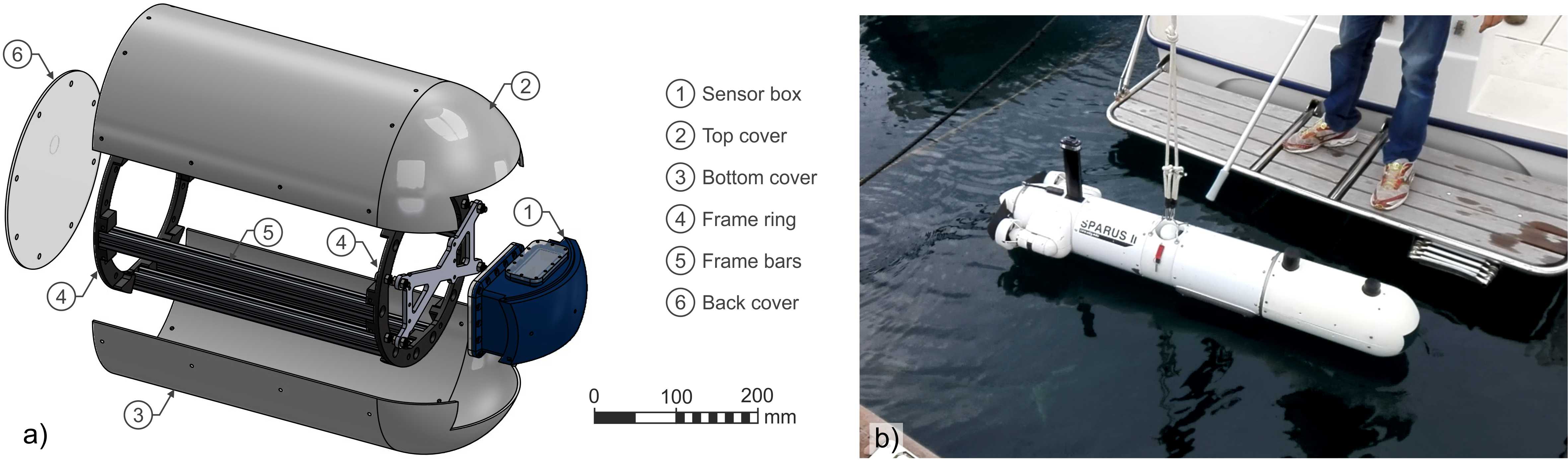}}
	\caption{Differential pressure sensor speedometry system for torpedo shape underwater vehicles based on ALL. a) Sketch of the design. b) Open ocean test with an underwater torpedo shape vehicle.}
	\label{Fig3_FlowSensing}
\end{figure*}

Considering the detected technological limitations and the knowledge gathered from previous steps, using an engineering approach, a new generation of ALL based in differential pressure sensor is designed, tested and validated (\cite{Fuentes-Perez2016}). These sensors systems demonstrate the ability to solve many of the problems of their predecessor (e.g. calibration, sensitivity or hydrostatic noise) as well as their utility and potential in environmental monitoring and underwater robotics. As an example of this technological advance the developed technology has been installed in a real underwater vehicle to estimate its surge velocity, demonstrating similar performances than the standard technologies (\cite{Fuentes-Perez2018})((Fig.~\ref{Fig3_FlowSensing}).

\section{Action Line Perception}
	
	\subsection{Marine Debris Detection with Neural Networks}
\textbf{Fellow}: Matias Valdenegro-Toro.\\
\textbf{Host Institution}: Heriot-Watt University, Edinburgh, United Kingdom.\\

Human activities are constantly polluting the environment. Marine environments are affected by floating or submerged human-made garbage, called marine debris. While the optional solution would be not to pollute marine environments, there is still the problem of cleaning up currently present debris. Autonomous Underwater Vehicles can be used to detect marine debris using a forward-looking sonar, in a similar way that underwater mines are detected.

Marine debris has a high variability in shape and possible views. Classic detection algorithms cannot model such variability.
We propose the use of neural networks as a general framework for different problems related to detecting marine debris.
We cover image classification, patch matching, and detection proposals/object detection.

\subsubsection{Sonar Image Classification}

Template matching techniques are state of the art for sonar images, but in the case of marine debris, these techniques perform poorly. In \cite{valdenegro2016object} we show that for our dataset of marine debris objects, cross-correlation template matching can obtain $93-98$ \% test accuracy, while a convolutional neural network can easily obtain over $98.2$ \% accuracy. In contrast we evaluated classic machine learning algorithms, namely support vector machines, gradient boosting, and random forests, which obtain $90-97.5$ \% accuracy. A convolutional neural network requires less parameters to obtain high accuracy.

Executing a trained neural network on embedded devices is not trivial, mostly due to the large number of required computations, which is directly linked to the number of trainable parameters. In \cite{valdenegro2017rtcnns} we developed neural networks based on ideas from SqueezeNet (\cite{iandola2016squeezenet}), which can achieve high accuracy with a low parameter count. We can obtain $98.8$ \% accuracy with only 1160 parameters, which contrasts to the 930K parameters required for larger networks. This leads to a $28.6$ speedup on a Raspberry Pi 2, where our network can recognize images at 42 milliseconds per frame (approx 24 frames per second).

\subsubsection{Matching Sonar Image Patches}

A big open problem in sonar perception is matching two sonar image patches. Given two patches, an algorithm must decide if they correspond to the same object/scene from different perspective. This is difficult due issues inherent to acoustic imaging, such as noise, viewpoint dependency, and shadows. In \cite{valdenegro2017improving} we show that a convolutional neural network can successfully learn to discriminate and match sonar patches with high accuracy, up to $91$ \% Area under the ROC Curve (AUCs). In contrast, classic keypoint matching like SURF, SIFT, and ORB, obtains only $61-63$ \% AUC, very close to random guessing. Classic Machine Learning methods like SVMs and Random Forests work slightly better, at $65-74$ \% AUC. Matching scores produced by a neural network are considerably more discriminative at matching sonar image patches, allowing for new applications such as SLAM and one-shot object detection/recognition.

\subsubsection{Detection Proposals}

\begin{figure}
	\centering
	\subfloat[FLS Image]{
		\includegraphics[width=0.15\textwidth]{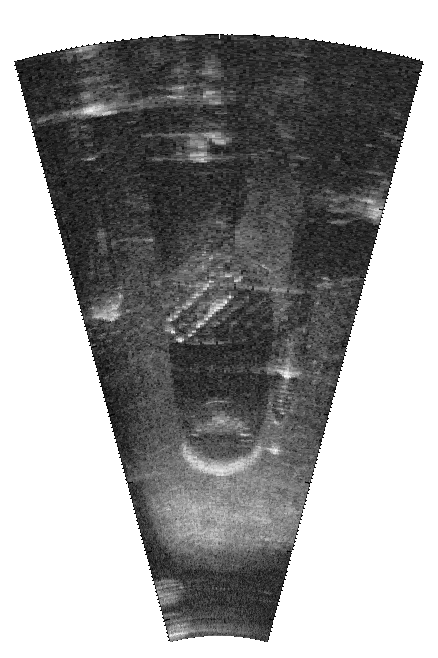}
	}
	\subfloat[Objectness]{
		\includegraphics[width=0.15\textwidth]{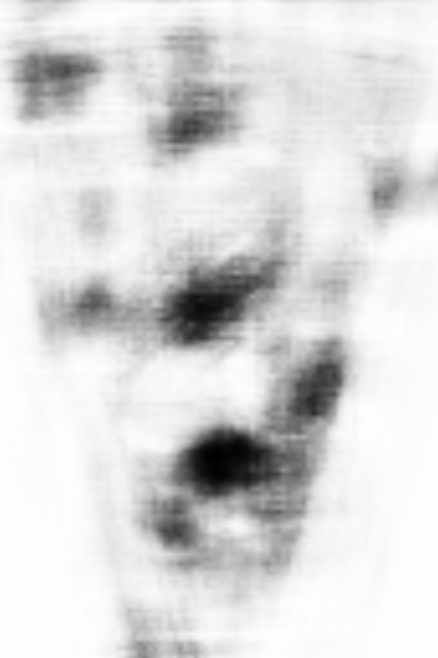}
	}
	\subfloat[Detections]{
		\includegraphics[width=0.15\textwidth]{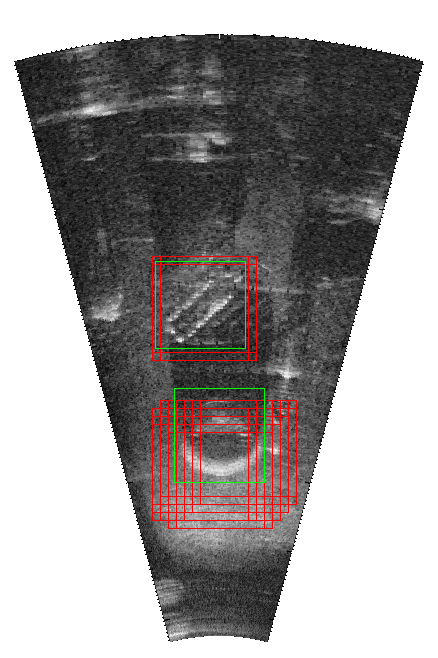}
	}
	\caption{Example objectness map and detections produced by our method. Light shades represent low objectness scores, while Dark shades represent high objectness score.}
	\label{objectnessDetections}
\end{figure}

Marine debris usually have complex shapes that cannot be considered a priori. Detecting these kind of objects requires detector algorithms that do not rely on shape or class information. A generic object detector is called a Detection proposals algorithm, and in \cite{valdenegro2016objectness} we have built a variation of such technique that is specifically designed for sonar images.

We train a convolutional neural network to predict a per-pixel value called "objectness", which quantifies the likelihood of an object being present. From these values generic object detections can be obtained by either thresholding, or ranking and finding maxima in the objectness map. A sample image with an objectness map and produced detections is shown in Fig. \ref{objectnessDetections}.

We evaluated this method on our marine debris dataset, where we obtain $93$ \% recall. In comparison, a template matching with cross-correlation method only obtains around $89$ \% recall. We also evaluated two methods from the state of the art in optical color images. EdgeBoxes obtains $78$ \% recall, while Selective Search obtains $86$ \%. While our method uses a single scale, it outperforms methods based on multiple scales.

\begin{figure}
	\centering
	\subfloat[FLS Image]{
		\includegraphics[width=0.15\textwidth]{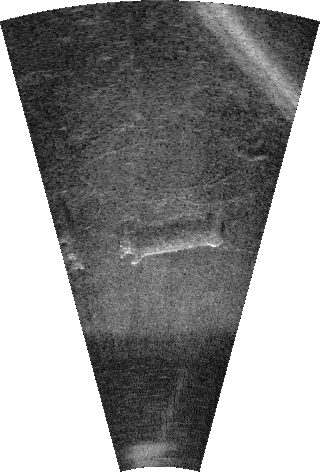}
	}
	\subfloat[Objectness]{
		\includegraphics[width=0.15\textwidth]{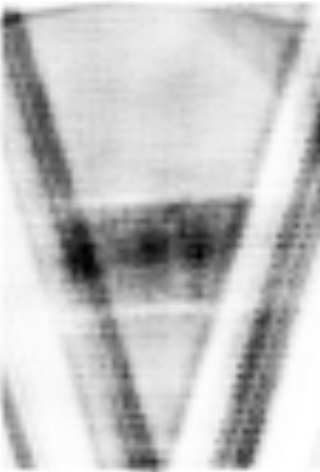}
	}
	\subfloat[Detections]{
		\includegraphics[width=0.15\textwidth]{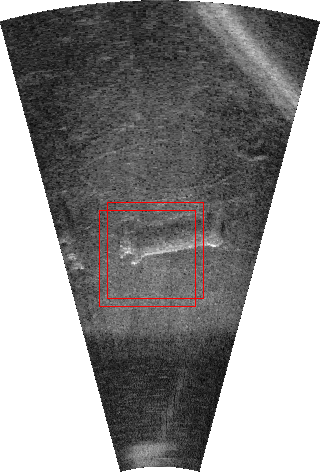}
	}

	\subfloat[FLS Image]{
		\includegraphics[width=0.15\textwidth]{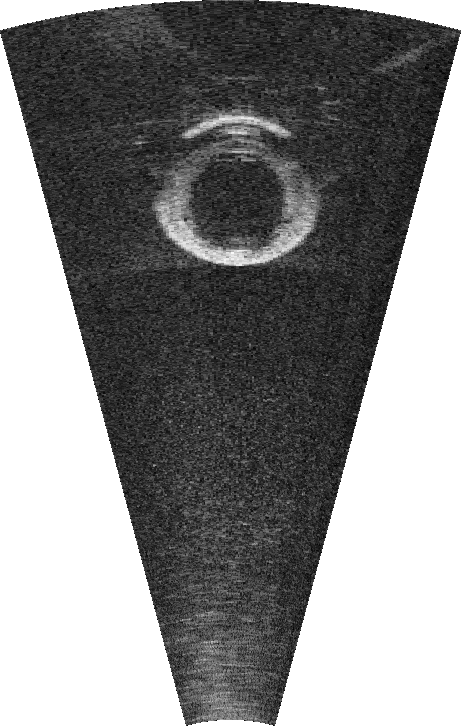}
	}
	\subfloat[Objectness]{
		\includegraphics[width=0.15\textwidth]{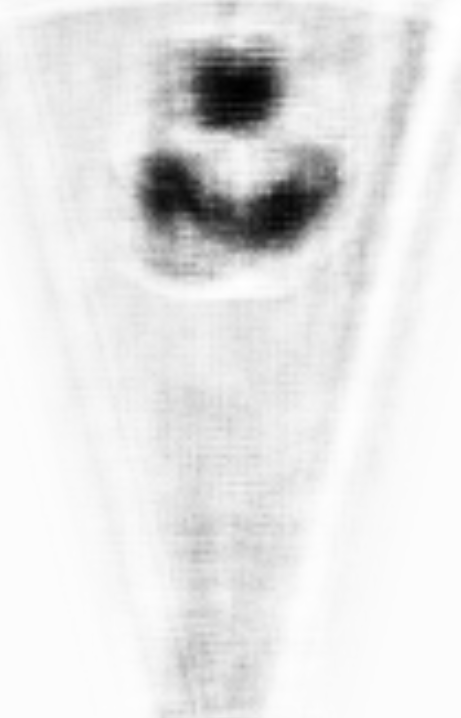}
	}
	\subfloat[Detections]{
		\includegraphics[width=0.15\textwidth]{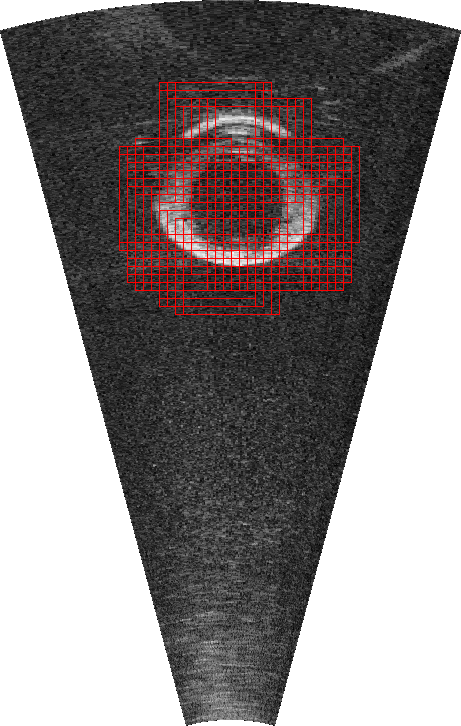}
	}
	\caption{Examples with unseen objects. The top row shows a Wrench, while the bottom row shows a large rubber tire. None of these objects are present in the training set.}
	\label{unseenDetections}
\end{figure}

Our method has the ability to adapt to different sonar devices. We have only evaluated it on forward-looking sonar data, but there should be no issues with modelling other sonar sensors like sidescan or synthetic aperture. Generalization is also very good, as the concept of an object in a sonar image generalizes to completely new objects, as shown in Fig. \ref{unseenDetections}.

\subsubsection{Object Detection}

We combine our previous detection proposals method to perform full object detection (\cite{valdenegro2016end}), by adding an additional output head to the network which outputs class probabilities. The advantage of this setup is that it can be trained end-to-end, meaning that only labeled bounding boxes with class information are needed to train the system, while the convolutional neural network will learn the appropriate features automatically.

Our method obtains $75$ \% classification accuracy with $93$ \% proposal recall. One way to improve these results is by using a SVM for classification on the features learned by the convolutional network. In this case proposal recall does not change, but classification accuracy improves to $85$ \%. In contrast, a template matching with cross-correlation obtains $89$ \% proposal recall with $65$ \% classification accuracy. Another method developed in \cite{valdenegro2016submerged} that uses only a classifier without objectness information obtains $71$ \% accuracy.	
	\subsection{Biosonar for Object Characterization} 
\textbf{Fellow}: Mariia Dmitrieva.\\
\textbf{Host Institution}: Heriot-Watt University, Edinburgh, United Kingdom.\\

Dolphins identify objects using their sonar, which works by emitting short acoustic pulses with high bandwidth and high intensity. They can evaluate object’s size, shape and material properties of the object, by processing echo reflected from the target. The study of the dolphins’ clicks inspired a simulation of the signals for echolocation purposes. They are already used for object characterisation in \cite{Pailhas:2010} \cite{Qiao:2016} \cite{Dmitrieva-LM:2015} \cite{Dmitrieva-MLSP:2017}. This research is focused on the object' material recognition using wideband pulses. This section presents a main contribution of the research, which includes material based classification and material recognition.

The wideband sonar works in the frequency range from $30 kHz$ to $160 kHz$ and allows transmitting pulses of different shape and duration. Linear down chirp pulses were used in this work. 
The objects in the study are limited by a 2-layer spherical shell with variety in radius, shell thickness, shell material and filler material.

\begin{figure}[!htb]
 \centering 
\begin{minipage}[h]{.32\linewidth}
  \centering 
  \centerline{\includegraphics[width=1\textwidth]{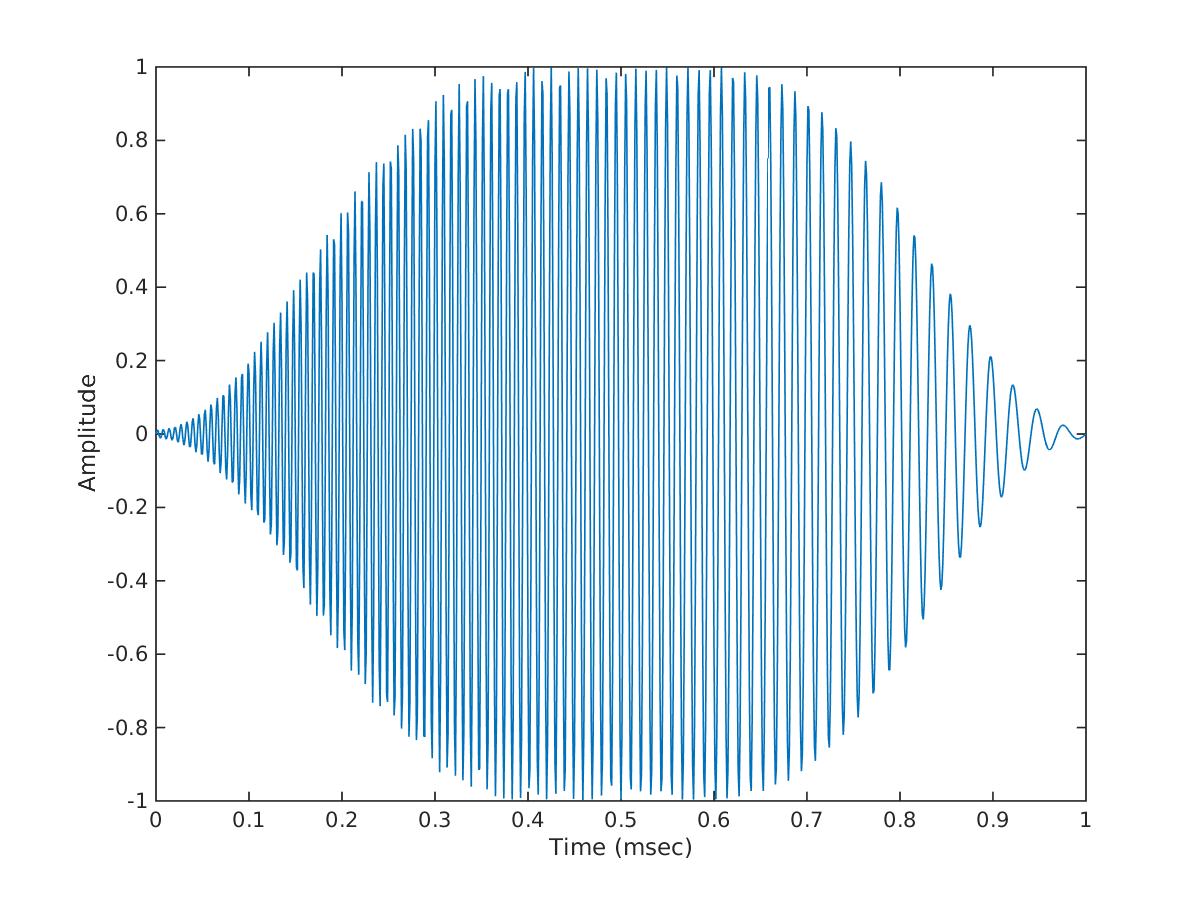}}
  \centerline{(a) TD}\medskip
\end{minipage}
\begin{minipage}[h]{0.32\linewidth}
  \centering
  \centerline{\includegraphics[width=1\textwidth]{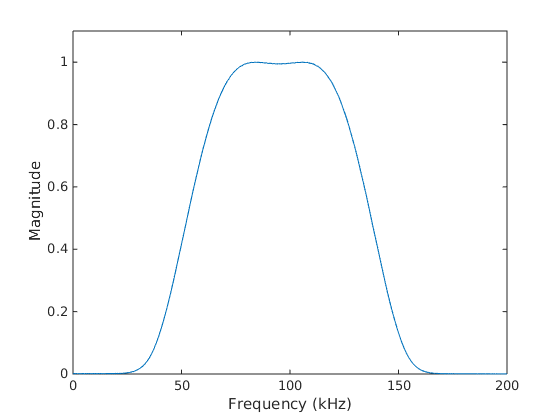}}
  \centerline{(b) FD}\medskip
\end{minipage} 
\begin{minipage}[h]{0.32\linewidth}
  \centering
  \centerline{\includegraphics[width=1\textwidth]{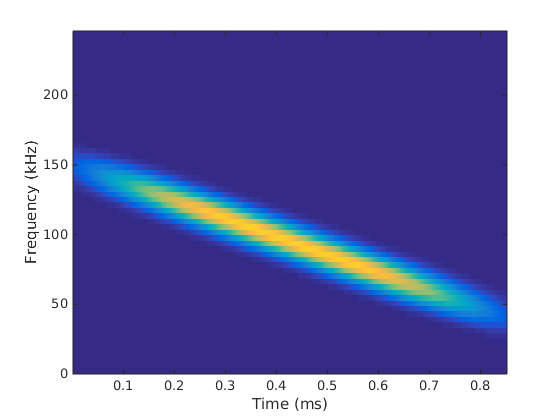}}
  \centerline{(c) TFD}\medskip
\end{minipage} \\
\begin{minipage}[h]{.32\linewidth}
  \centering 
  \centerline{\includegraphics[width=1\textwidth]{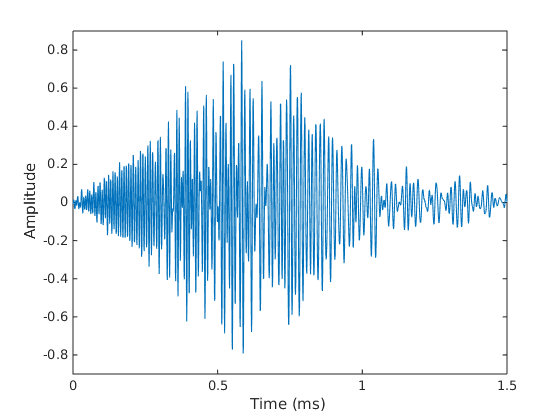}}
  \centerline{(d) TD}\medskip
\end{minipage}
\begin{minipage}[h]{0.32\linewidth}
  \centering
  \centerline{\includegraphics[width=1\textwidth]{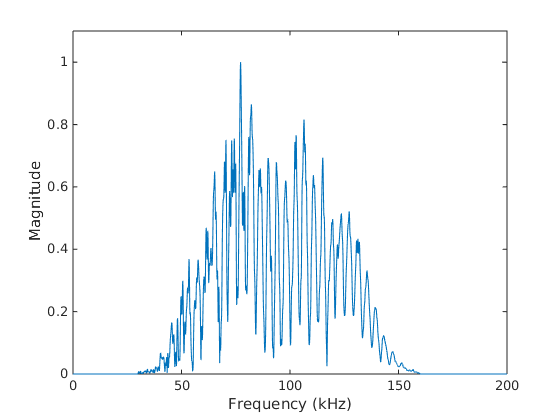}}
  \centerline{(e) FD}\medskip
\end{minipage}
\begin{minipage}[h]{0.32\linewidth}
  \centering
  \centerline{\includegraphics[width=1\textwidth]{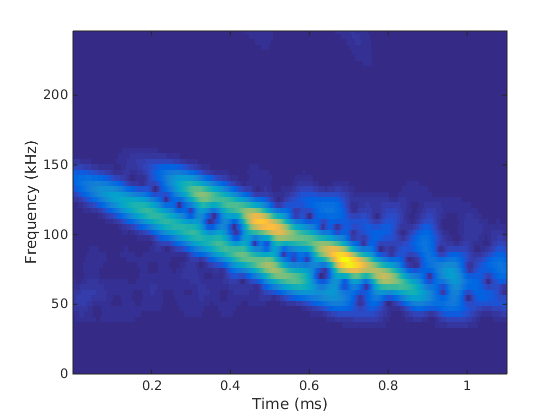}}
  \centerline{(f) TFD}\medskip
\end{minipage}
\caption{Initial ((a)-(c)) and reflected ((d)-(f)) pulses}  
\label{fig:reflection}
\end{figure}

The echo reflected from an object is changed due to the interaction of the sound wave with an object during the reflection process (outer shell and material structure)  \cite{Hickling:1962}. Figure \ref{fig:reflection} presents an initial pulse and an echo reflected from an aluminium sphere filled with water. The initial pulse is generated in the frequency range $(160-30)kHz$ and  the $3dB$ bandwidth of the pulse is $(136 - 52)kHz$.
These changes in the pulse are influenced by the shape of the object, its size, the shell thickness, filler and shell materials. Each sphere has its unique combination of these parameters, and hence it has a unique reflected echo, which represents the target. 
In this work, the changes in the composition of the wideband pulses are used for the material recognition. 

The first step towards the material recognition is an object classification  based on object's material \cite{Dmitrieva-UACE:2017}. The objects have the same spherical shape, but differ in size, material of the shell and filler, thickness of the walls. In \cite{Dmitrieva-UACE:2017}, objects are classified based on their filler material.
Two classes of objects are formed for the study: filled with water and filled with air.  The objects are described
by their form function. The form function expresses a pressure field scattered from a target in a range of frequencies. It describes the way an object scatters a pulse, which makes it a good descriptor of an object itself. Form function has distinguished peaks and notches related to the object properties. It is calculated from the reflected signal and handled as a descriptor for the classification with Multilayer Perceptron (MLP) and Support Vector Machine (SVM). The MLP has a five fully connected layers configuration with ReLU activation for all the layers except the output layer, which uses Sigmoid activation. SVM classifier is chosen with a Radial Basis Function kernel. The classification is performed with 3-Fold cross-validation for both classifiers to eliminate effect of data on the classification accuracy and provide fair evaluation of the results.
 The results for the form function descriptor is compared to the representation of the echo in frequency domain. The highest accuracy of $(98.8 \pm 1.31)\%$ is achieved with the form function descriptor and MLP classifier. The study illustrates possibility of object classification based on a single characteristic of the objects - filler material.

 The machine learning solution requires a training data. It also cannot be used to distinguish between materials with similar properties (water and salt water). Another solution was proposed to recognise material of the shell and the filler, which is based on the timing of the reflected components. The reflected echoes are formed by a number of processes which occur during the reflection of the initial pulse from the object. The components include specular reflection, diffracted circumferential waves, elastic waves transmitted into the shell and bouncing from the inner shell surface, Lamb waves and others  \cite{Guillermo:1998}.  The timing of the reflection components and geometry of the waves’ paths determine speed of sound in the mediums where the wave is propagating. The components of the echo are detected using a matched filtering technique and a peak detector. This material identification approach doesn’t include any machine leaning techniques and doesn’t require training data. Material identification is based on the information about the cylinder’s radius, the outer medium and recording of the reflected echo. The approach was tested for synthetic and experimental data. The data includes objects with variety in radius, shell thickness, shell material and filler material.

This work has provided a solution for the material based classification and material identification using wideband pulses. The solution is limited by a spherical object. The future research will be focused on the extension of the solution to a cylinder shape and work on the adaptive signals.
	\subsection{Optical Mapping and Disturbance Rejection} 
\textbf{Fellow}: Klemen Istenic.\\
\textbf{Host Institution}: University of Girona, Girona, Spain.\\

The ability to accurately map underwater environments is becoming increasingly important in modern marine surveys. Resulting 3D models convey immense information easily interpretable by humans and enable experts to perform further in-depth investigations of areas of interest after the mission is conducted. Furthermore, with advances in UUV technologies, scientists are able to access large marine areas and deep sea regions and perform studies in interactions among physical, chemical and biological components of the environment. To correctly identify and correlate changes with possible underlying factors, long-term and high-frequency observations of the studied ecosystem are required, as well as systematic evaluations of the quality of produced models in each of the surveys.

Underwater 3D mapping can be performed using either acoustic (\textit{e.g.} multibeam echosounders and sidescan sonars) or optical devices (cameras). While elevation maps produced with acoustic sensors are indispensable for providing a rough approximation of the terrain, their relatively coarse resolution limits its usability for highly detailed representations of complex structures with concavities, often desired by scientists. Optical sensing, on the other hand, can be used to recover high resolution 3D representations of smaller areas of interest, but requires close proximity acquisitions due to the effects of light attenuation and scattering processes in the water medium. For this reason, optical surveys are performed either using an ROV, operated by a human pilot, or by an AUV conducting a mission in a previously explored area, enabling the vehicle to navigate at a constant and safe distance from underwater structures and seafloor. Lack of preliminary information about the scene therefore requires the usage of an ROV which relies on (often expensive) surface vessels.

\begin{figure}[!htbp]
	\begin{center}
		\includegraphics[width=3.3in]{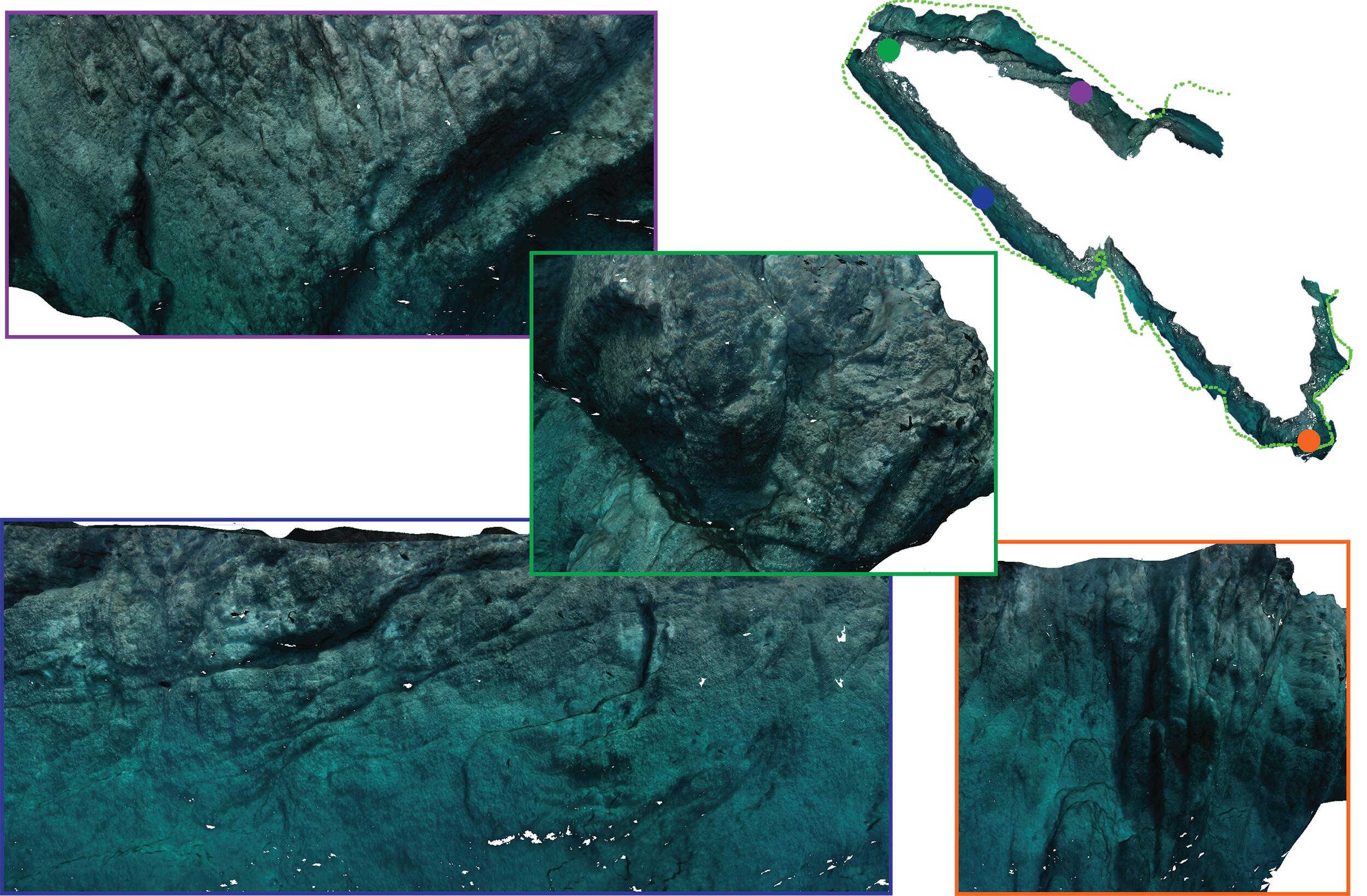}
	\end{center}
	\caption{ Example of optical 3D reconstruction of an explored environment (top view with a few enlarged details)}\label{optical3dRecon}
\end{figure}

In~\cite{hernandez2016autonomousA} we presented a framework that endows an AUV with the capability to autonomously navigate and gather optical data in environments for which no previous information is available, and an optical based 3D reconstruction pipeline for producing 3D models. The collision-free paths are computed taking into the consideration range constraints of optical data acquisition and the data collected, using an arbitrary camera setup, is subsequently processed to produce various 3D representations (i.e., sparse, dense, meshed, textured) of the surveyed area. The developed framework has been used on a Sparus II (University of Girona, Girona, Spain) AUV in multiple challenging real-world and natural scenarios (\cite{hernandez2016autonomousB,vidal2017online}) with an example of textured model shown in~\ref{optical3dRecon}.

Given the need that marine scientists have in  using 3D models as base maps for scientific studies, it becomes very important to study and characterize the influence of conditions and strategies undertaken during the acquisition process on the quality of the final model. Furthermore, the ability to estimate the uncertainty of the model during the mission, would enable concurrent assessment of the quality of the acquired data as well as the identification of poorly mapped or even missing areas. This can be then further exploited by skilled pilots and autonomous planning schemes, to guide the vehicle towards problematic areas and collect additional data to significantly improve the final mapping result and thus the mission efficiency.

While techniques, such as simultaneous localization and mapping (SLAM) and light bundle adjustment can produce real-time sparse 3D reconstructions, neither can provide an estimate of the uncertainty of the solution encoded in the covariance matrix. On the other hand, structure from motion (SfM) and bundle adjustment (BA) techniques performed on a complete set of data are computationally expensive for large scale applications and are normally performed off-line. Such decoupling of the two steps prevents any automatic feedback about the quality of the reconstruction during the mission and consequently require careful mission planning and strong human intervention during the survey.

To address this issues, we proposed a novel and efficient incremental BA technique~\cite{ila2017fast}, which does not rely on approximations or partial and windowed optimizations, but achieves high performance by automatically restricting the optimization to only the set of variables that need to be optimized. The proposed technique is especially efficient for long camera trajectory applications where images are processed sequentially and points quickly move in and, soon after, out of the field of view. Additionally, the solution also provide feedback about the quality of the reconstruction (marginal covariances), which is on average one order of magnitude (almost two in some cases) faster than the high-end solvers. 

\begin{figure}[!htbp]
	\begin{center}
		\includegraphics[width=3.3in]{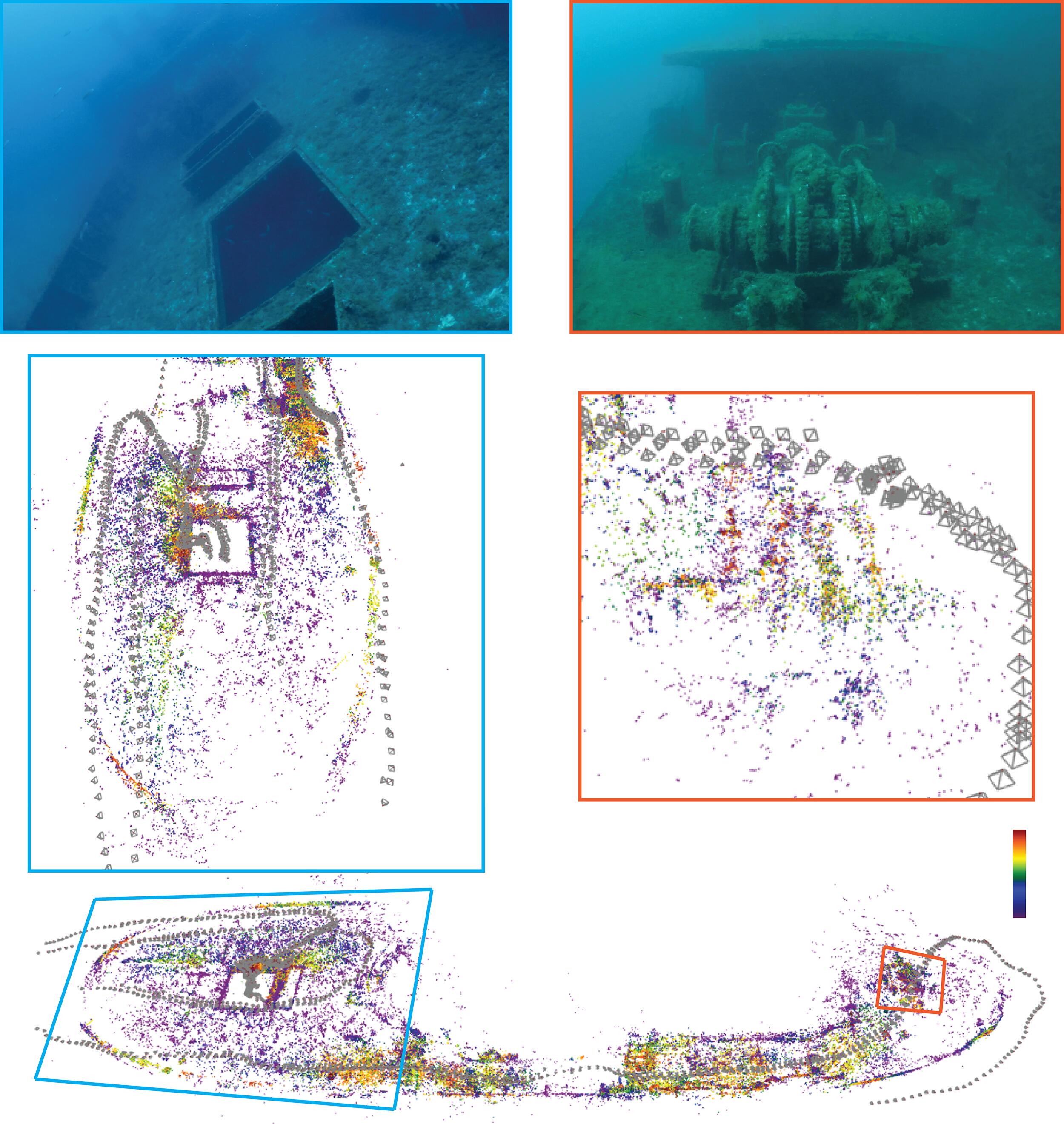}
	\end{center}
	\caption{Sparse 3D reconstruction with color-coded magnitude of the uncertainty estimation (violet-high uncertainty, red-low uncertainty)} \label{optical3dQuality}
\end{figure}

With the aim of producing robust, globally-consistent and large-scale underwater 3D reconstructions, the incremental BA solver has also been incorporated into a new SfM pipeline proposed in~\cite{istenic2017mission}. Due to the nature of the water medium, a new two-layered mapping approach was introduced to reduce the effects of plausible outliers on global reconstruction, while ensuring a robust and continuous reconstruction. As such, the new system can provide not only the solution of the optimization (camera trajectory along time and the 3D points of the environment), but also the estimate of the uncertainty associated with the 3D reconstruction, as seen on figure~\ref{optical3dQuality} (with uncertainty of the scene depicted using a color scheme). Results are obtained in mission-time, \textit{i.e.} while the robot is in the water or very shortly afterwards, enabling before-mentioned re-planning of the missions. 

\begin{figure}[!htbp]
	\begin{center}
		\includegraphics[width=3.3in]{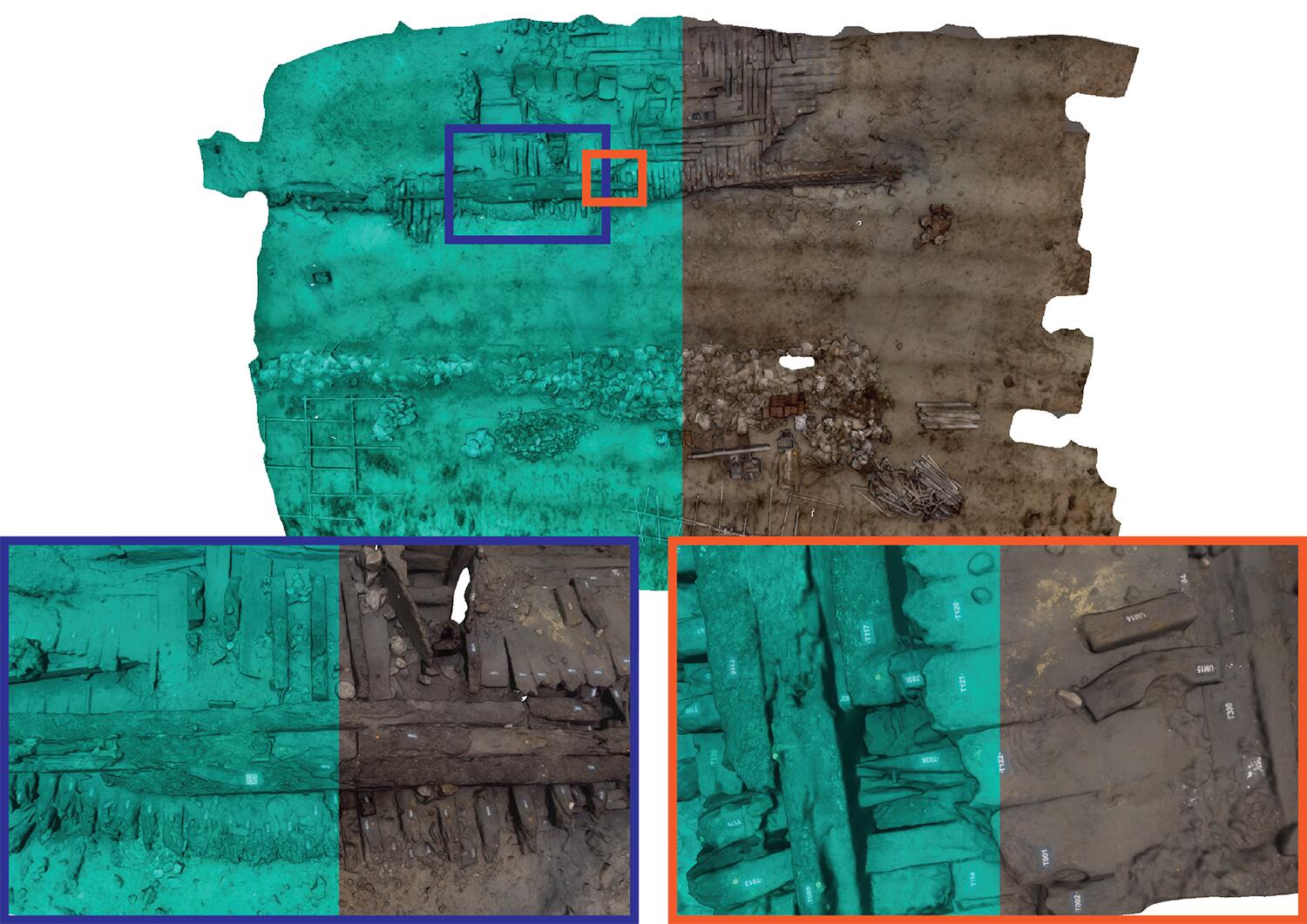}
	\end{center}
	\caption{Textured 3D model before and after color correction} \label{optical3dColor}
\end{figure}

In our attempt to produce 3D models for scientific use, we did not solely focus on the geometric aspects but also attempted to mitigate the effects of wavelength dependent light attenuation in the water. The ability to reproduce models with both geometric and photometric accuracy will provide scientists with another dimension of valuable information about the environment. In a currently unpublished work, we exploited the knowledge about the geometry of the scene (obtained by reconstructing the 3D model) to estimate and reverse the effects of the medium. An example of a model before and after the color recovery is shown in figure~\ref{optical3dColor}.

	\subsection{3D reconstruction and object recognition from 2D SONARs}
\textbf{Fellow}: Thomas Guerneve.\\
\textbf{Host Institution}: SeeByte, Edinburgh, United Kingdom. \\

While optical mapping can provide detailed 3D information, they typically require pristine visibility conditions that are only met in clear water areas and at short observation ranges (1-3m). On the opposite, acoustic sensing modalities provide lower spatial resolutions (typically 1 to 5cm) but enable sensing in low-visibility conditions at ranges of up to 100m. 2D SONARs, often referred to as multibeam SONARs provide direct range and bearing angle readings but the integration along the elevation angle occurring during the imaging process make the recovery of the 3D structure of the scene an ill-posed problem. When operating in low-visibility conditions, it is of interest to leverage the sensing capability of SONAR sensors to obtain an accurate 3D representation of the environment to allow for 3D mapping, path planning, collision avoidance and object recognition.\par
In order to address the problem of 3D reconstruction from 2D SONARs, we proposed two novel reconstruction techniques based on the association of multiple images acquired along the direction of uncertainty of the sensor (elevation angle). The first method, presented in \cite{guerneve2015carving} is an iterative method enabling per-sample reconstruction based on the principle of space carving. On the contrary the second reconstruction technique presented in {DOI:10.1002/rob.21783} requires to have all samples available at the time of the reconstruction (batch processing) and is based on a blind deconvolution formulation. A model-to-data constrained optimization approach enabled to perform the deconvolution. We performed the first and extensive quantitative analysis of 3D reconstruction using 2D SONARs. Importantly, our work showed the interest of using a wide-beam sensor over so-called pencil-beams sensors to increase the surface coverage of underwater observations. The study compared both methods and exhibited improved 3D reconstruction accuracy when using the deconvolution method over the space carving method. Both methods showed to be suited for on-field AUV operations and have different hardware requirements. Due to the maintenance of a 3D map, the space carving method requires large amount of RAM (up to 16GB) but the processing part does not require any intensive CPU usage. On the contrary, the deconvolution method has a small memory footprint (200MB at most) but requires more intense processing due to the Non-Negative Least Square optimization.\par
\begin{figure}[htbp]
\centering
\begin{tabular}{c c c}
\includegraphics[width=0.29\linewidth]{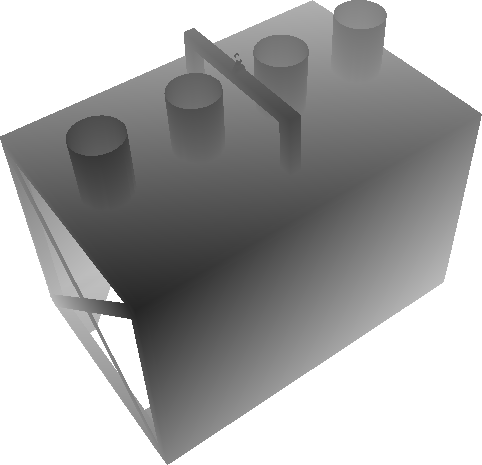} &
\includegraphics[width=0.29\linewidth]{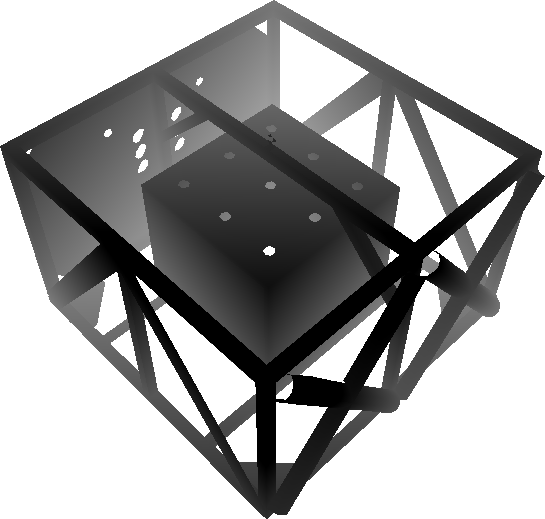} &
\includegraphics[width=0.29\linewidth]{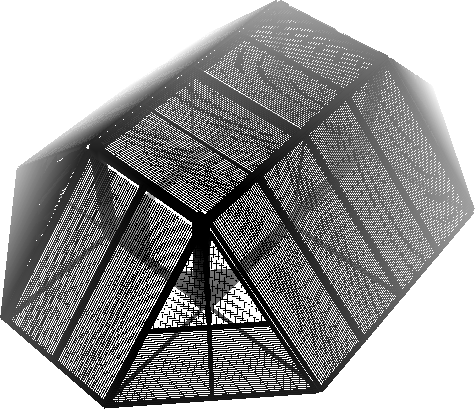}\\
(a) & (b) & (c) \\
\includegraphics[width=0.29\linewidth]{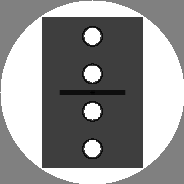} &
\includegraphics[width=0.29\linewidth]{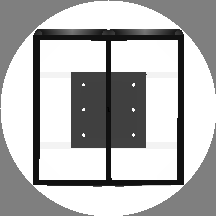} &
\includegraphics[width=0.29\linewidth]{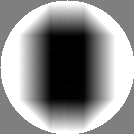}\\
(d) & (e) & (f) \\
\includegraphics[width=0.29\linewidth]{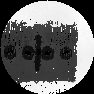} &
\includegraphics[width=0.29\linewidth]{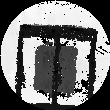} &
\includegraphics[width=0.29\linewidth]{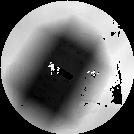}\\
(g) & (h) & (i) \\
\includegraphics[width=0.29\linewidth]{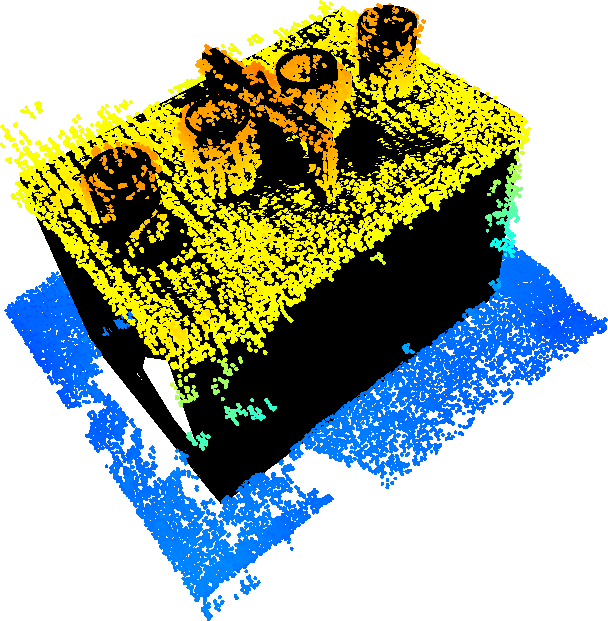} &
\includegraphics[width=0.29\linewidth]{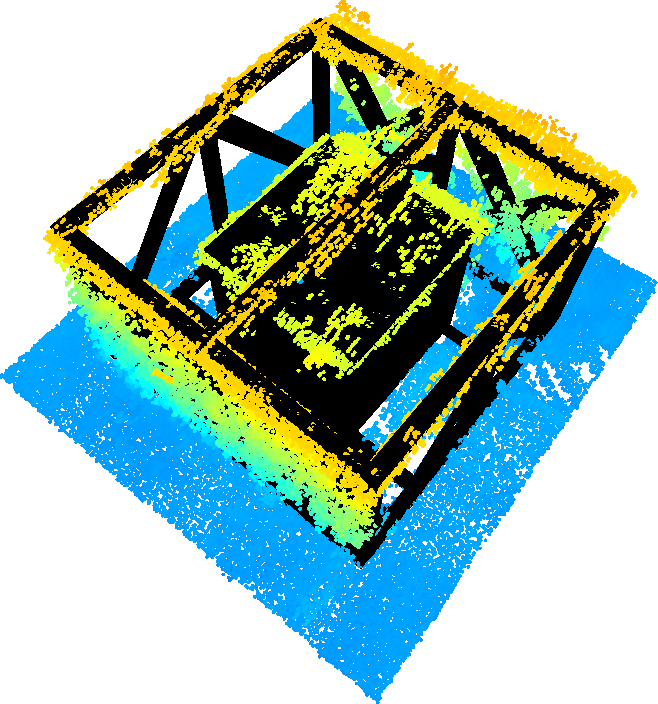} &
\includegraphics[width=0.29\linewidth]{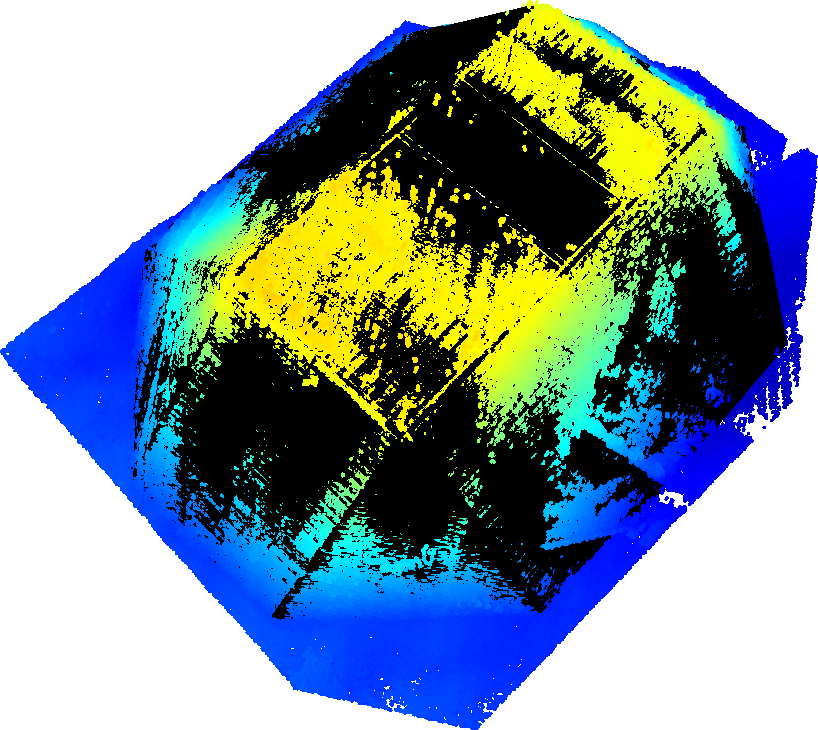}\\
(j) & (k) & (l) \\
\end{tabular}
\caption{CAD-model-based structure recognition. a-b-c) The CAD models used as prior information are converted to elevation maps (d-e-f) to perform a structure recognition step on the full field, leading to the selection of the most similar patches containing the structures of interest (g-h-i). The CAD model of each structure is then finally registered to the scene (j-k-l).}
\label{fig:recognitionThomas}
\end{figure}

On a more general point of view, the 3D reconstruction accuracy is limited by the accuracy of the SONAR (centimeter level) as well as the navigation of the platform (typically a couple of centimeters drift over a few meters). Based on this observation, our work presented in \cite{guerneve2017IROS} focused on demonstrating that 3D from SONAR was suitable for underwater object recognition. In this study, a 50x34 metres field was mapped using a downward-looking multibeam SONAR mounted on an AUV. A bundle adjustment process enabled to maintain geometrical consistency between consecutive swathes and reduced navigation drift to a few centimetres only. Once the map of the field obtained, a histogram-based description was performed on a set of circular stencils enabling rotation invariant description of the scene. As illustrated in fig.\ref{fig:recognitionThomas} set of rough CAD models of the structures of interest was used and compared to the description of the scene to perform structure recognition and localization in a small amount of time. After recognition, a two-steps matching process was applied and enabled accurate registration of the CAD models with median registration errors typically under 3 centimetres. More importantly, this work showed the possibility of generating 3D models of underwater environments solely based on a multibeam SONAR and a set of CAD models. A particularly important application of this study is the deployment of AUV in unknown or partially unknown environments such as offshore fields where only partial information is provided (high uncertainty on the appearance and location of the structures of interest). In this context, the safe operation of autonomous underwater vehicles such as the Subsea7 AIV requires an accurate 3D model of the world, making our work relevant to this project.
\par
Following the demonstration of structure identification and localization on an underwater field, we investigated a CAD-model-based video mapping approach, enabling to leverage the high-resolution of CAD models. Similarly to our previous work, we presented in \cite{guerneve2017cad} the robust registration of the CAD model in the scene enabled to obtain an outlier-free 3D representation of the world from the 3D reconstruction of the field. A simple raytracing approach then provided a coloured 3D representation of the structure. A natural and practical application of this work is the video inspection of underwater features which typically requires analysis of a large amount of video data (i.e. multiple sequences acquired around the object of interest). In this context, the ability to generate realistic coloured 3D models is highly desired and offers an instant visual summary of the state of underwater equipments.

\section{Lessons Learned}

\begin{enumerate}
    \item The lack of public datasets in marine and underwater applications limits the use of data-driven techniques, such as Machine Learning.
    \item Due to the previous point, capturing and labeling data is potentially a large part of PhD research, limiting work in other areas.
    \item Sensors inspired in biological structures are a viable approach for successful sensing in underwater environments. This is clearly shown by the Biosonar and the Artificial Lateral Line.
    \item Modeling the behavior of AUV components require large training sets, and there is no methodology on how to efficiently capture such data.
    \item Combining multiple views of a given scene can produce unexpected results due to noise reduction and averaging, which is useful for perception in adversarial environments such as underwater.
\end{enumerate}

\section{Conclusions}

This work represents a short summary of the scientific work performed by the Robocademy Marie Sk\l{}odowska-Curie ITN fellows during the duration of the three year project. 71 publications in top venues were produced, with at least 10 more in the publication pipeline. We believe this shows the success of the project in numerical terms, but the final contribution of this European investment will be the fellows that have become scientific experts while pursuing Doctoral degrees in Underwater Robotics.

We expect that our research results can be quickly transferred to Industry. For example Graaltech is using Robocademy research to survey submarine oil wells, and SeeByte will use our 3D reconstruction techniques from sonar. Still AUVs lag behind other kinds of Robots, and more research and development is needed to improve their capabilities.
	
\begin{ack}
	This work has been partially supported by the FP7-PEOPLE-2013-ITN project ROBOCADEMY (Ref 608096) funded by the European
	Commission. Three fellows did not contribute to this paper but were still part of the network: Samy Nascimento, Diogo Machado, and Youssef Essaouari.
\end{ack}
	
\bibliography{biblio}             
	
\end{document}